\documentclass[11pt]{article}

\usepackage{acl}

\usepackage{bm}

\usepackage{times}
\usepackage{latexsym}
\usepackage{amsmath}
\usepackage{booktabs}
\usepackage{multirow}
\usepackage{tabularx}
\usepackage{array}
\usepackage{soul}
\usepackage{subcaption}
\usepackage{algorithm}
\usepackage{algpseudocode}
\usepackage{paralist}
\usepackage{graphicx}
\usepackage[table]{xcolor}
\usepackage{colortbl}
\newcolumntype{C}{>{\centering\arraybackslash}X}
\newcolumntype{L}{>{\raggedright\arraybackslash}X}

\usepackage[T1]{fontenc}

\usepackage[utf8]{inputenc}

\usepackage{microtype}

\usepackage{inconsolata}
\usepackage{xcolor}

\usepackage{graphicx}
\usepackage{amssymb}
\usepackage{float}
\newcommand{\method}[1]{\textsc{DMDIntel}}
\newcommand{\dmd}[1]{\textsc{DMD}}
\newcommand{\hdmd}[1]{\textsc{HoDMD}}
\newcommand{\ig}[1]{\textsc{IG}}
\newcommand{\pca}[1]{\textsc{PCA}}
\newcommand{\shap}[1]{\textsc{SHAP}}
\newcommand{\lama}[1]{\texttt{Llama-3.2-3B-inst}}
\newcommand{\qwn}[1]{\texttt{Qwen3-4B-inst}}
\newcommand{\mist}[1]{\texttt{Mistral-7B-v0.3-inst}}
\newcommand{\fk}[1]{\textbf{FakeEdit}}
\newcommand{\snt}[1]{\textbf{Sentiment}}
\newcommand{\hx}[1]{\textbf{HateXplain}}
\definecolor{fg}{RGB}{34,180,34}
\definecolor{lg}{rgb}{0.68, 0.87, 0.68}

\title{\method{}: Interpreting Large Language Models\\ via Dynamic Mode Decomposition}

\author{Amogh Joshi \\
  IIT Kharagpur, India \\
  University of Manchester, UK \\ \And
  Animesh Mukherjee \\
  IIT Kharagpur, India\\ \And
  Sergey Utyuzhnikov \\
  University of Manchester, UK \\
}

\begin{document}
\maketitle
\begin{abstract}

In this work, we introduce \method{} which uses \textit{dynamic mode decomposition} (\dmd{}) to make the predictions made by LLMs in a classification task interpretable. It develops an input attribution pipeline, that first decomposes the hidden states of an LLM into prominent patterns, also known as \textit{modes}, and then associates ranks to the input tokens based on the projection values on those modes. Rigorous experiments across three datasets and three model families consistently show that the ranked attribution of input tokens obtained using \method{} by far outperforms state-of-the-art techniques such as principal component analysis, integrated gradients and \shap{}.

\end{abstract}

\vspace{0.1cm}

\section{Introduction}
Transformer language models \cite{vaswani2017attention} redefined the processing of natural language by introducing self-attention. Following this, many modifications have been made to the transformer architecture, such as encoder-only and decoder-only models \cite{devlin2019bert, radford2018improving}. Due to an increase in the complexity of these models, interpretability remains a huge challenge. A significant body of literature has been dedicated to dissecting and interpreting the internal mechanisms of attention layers in transformer-based architectures.
Traditional interpretability methods, such as Integrated Gradients (\ig{}) \cite{sundararajan2017axiomatic}, \shap{} \cite{lundberg2017unified}, etc., are often used for input attribution in LLMs. Although these methods are effective in identifying ``what'' the model looks at, they treat the tokens as discrete entities and often overlook the sequential evolution of the latent information, providing a fragmented snapshot rather than a continuous narrative of model's reasoning.\\
\noindent In addition to attribution, there are some attempts to decode the structural logic of the MLP component of the decoder layer. \citet{geva2021transformer} suggests that the MLP module acts as a key-value memory bank, where the first linear layer is a pattern detector (keys) and the second is a value provider that updates the residual stream. \citet{hernandez2023linearity} states that much of the black-box computation within an MLP can be approximated via linear operations in high-dimensional space. However, even these structural insights frequently treat each token's hidden state as an isolated vector, failing to capture the logic that builds across a sequence.\\
\noindent To address these issues, in this paper, we attempt to take advantage of masked self-attention, which invokes the sequential nature of decoder-only LLMs by analyzing hidden states as they evolve token by token. By viewing these hidden states as trajectories, we can treat the supervised fine-tuned LLM as a dynamical system where the internal processing resembles a flow in the embedding space. We use Dynamic Mode Decomposition (\dmd{}) \cite{schmid2010dynamic} to find a surrogate linear operator, $\mathbf{A}$, which approximates the complex model underlying $\mathcal{F}$. Specifically, we model the state transition, $h_{t+1} = \mathcal{F}(w_{\leq t})$, using a linear relationship $h_{t+1}=\mathbf{A}h_t$, where $w_{\leq t}$ represents the tokens until the current token index $t$ and $h_t$ represents the hidden state of the model at the token index $t$. This linear operator is then decomposed using \dmd{} into interpretable spatio-temporal modes. These modes can be thought of as the low-dimensional structures that manifest when a model is supervised fine-tuned for a particular downstream task. We then use the dominant modes for input attribution to identify the most influential tokens. Thus, we can think of these tokens as the main drivers of the model's input processing, as it prepares the output by reading the input tokens.

\noindent \textbf{Key contributions}: Our key contributions in this work are as follows.
\begin{compactitem}
    \item We introduce \method{}, an input attribution framework for supervised fine-tuned LLMs that extracts the primary tokens driving a given output classification. We show that the hidden states of a supervised fine-tuned model, obtained from the \texttt{down\_proj} of its multi-layer perceptron (MLP) component, show interpretable patterns when decomposed into lower-dimensional spatio-temporal modes. The ranking of tokens by their projections onto these \dmd{} modes demonstrates that certain modes serve as robust indicators of the model’s information flow.
    \item We validate the versatility and robustness of our approach by evaluating it across three distinct LLM families of varying parameter scales using three diverse text classification datasets.
    \item To the best of our knowledge, this is the first work to analyze the sequential evolution of hidden states in decoder-only LLMs as a dynamical system for input attribution. We demonstrate that \method{} consistently identifies the highest proportion of ground-truth tokens while preserving their ranked importance, outperforming baseline methods such as \pca{}, \ig{}, and \shap{}. 
\end{compactitem}

\section{Background}
Dynamic Mode Decomposition (\dmd{}) \cite{schmid2010dynamic} was originally developed in fluid mechanics to extract coherent spatial structures from complex, non-linear flow fields. \dmd{} is a purely data-driven, model-agnostic technique that requires no prior knowledge of the underlying system governing equations. By operating directly on time-series observation snapshots, \dmd{} constructs a best-fit linear operator that approximates the non-linear dynamics of the system. The resulting eigendecomposition yields a set of spatial \dmd{} modes—representing the primary spatial structures along with complex eigenvalues that quantify their temporal growth, decay, and oscillation frequencies. A complete mathematical derivation and algorithmic setup of \dmd{} is detailed in Appendix \ref{app:dmd}.

\section{Related work}

\noindent\textbf{Interpretability of LLMs}: Various analytical frameworks have been proposed to understand the inner workings of LLMs. Probing-based approaches evaluate whether hidden representations capture structural or semantic features by training linear classifiers on top of frozen model layers \cite{tenney-etal-2019-bert}. Beyond static probing, mechanistic interpretability seeks to reverse-engineer specific network circuits, mapping exact weights and attention head paths to functional behaviors such as in-context pattern matching \cite{olsson2022context} or factual retrieval \cite{meng2022locating}. To observe how these internal features evolve into final predictions, methods such as the LogitLens decode intermediate hidden states directly into the vocabulary space. 
This concept is further refined in TunedLens \cite{belrose2023eliciting}, which trains linear adapters at each layer to more accurately map representation dynamics to the model's predictive trajectory.

\noindent\textbf{\dmd{} applications}: Beyond fluid dynamics, \dmd{} has been successfully applied to video processing scenarios. \citet{kutz2017dynamic} established \dmd{}'s capability for background modeling in video stream data, effectively separating the foreground from the background. \citet{erichson2019compressed} utilized a memory-efficient variant of \dmd{} to achieve foreground-background separation in streaming video data.

\noindent\textbf{\dmd{} for NLP tasks}:
In recent works, \dmd{} has also been used for feature extraction to improve performance in classification tasks. For example, \cite{sachin2019dynamic} uses \dmd{} to extract spatio-temporal features from text representations to capture the evolving sentiment trajectory across a sentence. Similarly, this dynamical perspective has been successfully applied to multimodal and spoken language tasks; for example, \cite{mao2020eigenemo} utilizes \dmd{} to derive spectral representations of audio utterances. 
\citet{vyshnav2020offensive} evaluates \dmd{} along with traditional sequence models to isolate consistent and underlying semantic patterns from highly noisy and user-generated social media text. Together, these works highlight the efficacy of \dmd{} in capturing non-linear dynamics across diverse linguistic modalities.

\setlength{\belowdisplayskip}{1pt} \setlength{\belowdisplayshortskip}{1pt}
\setlength{\abovedisplayskip}{1pt} \setlength{\abovedisplayshortskip}{1pt}

\section{The \method{} framework}

In this section, we outline the \method{} framework which we propose for interpreting the predictions of decoder-only LLMs when they are used to perform various text classification tasks. We view the LLMs as a discrete-time dynamical system, where the evolution of hidden states acts as a first-order approximation of the LLM $\mathcal{F}$, such that $h_{k+1} \approx \mathcal{F} (h_k)$. Subsequently, we use \dmd{} (see Appendix~\ref{app:dmd} for more details) to approximate $\mathcal{F}$, and finally decompose it to get a set of interpretable modes. 

\subsection{Formulation of the data matrix}

As an LLM processes a text sequence, it stores the information in the intermediate hidden states such that $h_{t+1} = \mathcal{F}(w_{\leq t})$. The information can be considered as the combined representation of the instruction prompt plus the information from the actual input sentence in some non-linear fashion. This information is passed through all the layers, before passing through the unembedding matrix, followed by softmax to generate the actual token. 
We target a layer, and in that layer, we consider outputs from the \texttt{down\_proj} of the multilayer perceptron. For an input sequence consisting of $w_1, w_2, ..., w_n$ tokenized words, we collect the vectors from the selected layer to form the matrix $\textbf{X}$ such that

\begin{equation}
\small
\textbf{X} = 
\begin{bmatrix}
\vert & \vert &        & \vert \\
h_1   & h_2   & \cdots & h_{n} \\
\vert & \vert &        & \vert
\end{bmatrix}
\label{orig_matrix}
\end{equation}
where each $h_i = \mathcal{F}(w_{\leq i})$.

\subsection{Noise removal}

We find that the instruction fine-tuned models encode a significant representation of the prompt template used during fine-tuning within the hidden representations of each token.  As \citet{hemati2017biasing} proves the sensitivity of \dmd{} to noise in the data, we denoise the data matrix obtained in Eq. \ref{orig_matrix} by removing the noise inflicted by the prompt template. For this, we pass an empty prompt consisting of the same exact system and user prompts without the input sentence, and then collect the hidden representation from the same layer from where we pick the token representations. We call this instruction \textit{bias}, $\bm{b}$. This \textit{bias} is then subtracted from each $h_i$ to obtain the debiased estimate of the hidden representation of the actual token $w_i$.

\begin{equation}
    \hat{h}_i = h_i - \bm{b}
\end{equation}
Thus, our resulting data matrix is the collection of the debiased hidden states of each token.

\begin{equation}
\small
\boldsymbol{\hat{X}} = 
\begin{bmatrix}
\vert & \vert &        & \vert \\
\hat{h}_1   & \hat{h}_2   & \cdots & \hat{h}_n \\
\vert & \vert &        & \vert
\end{bmatrix}
\label{debiased_matrix}
\end{equation}

\begin{figure*}[t]
    \centering
    \begin{subfigure}[b]{0.30\textwidth}
        \centering
        \includegraphics[width=\textwidth]{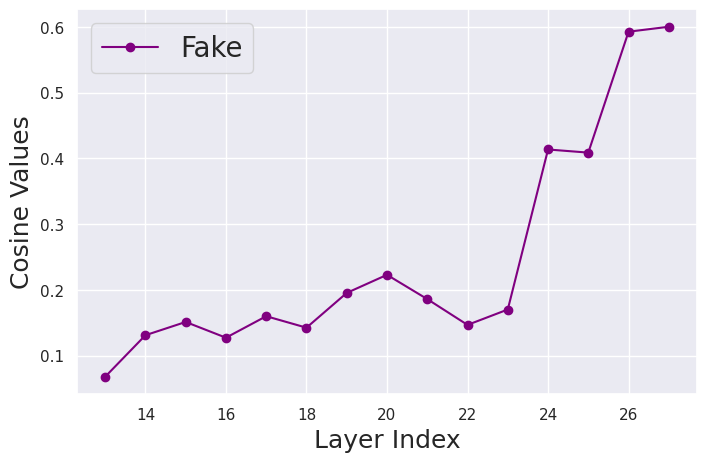}
        \caption{\lama{}: FE}
    \end{subfigure}
    \hfill
    \begin{subfigure}[b]{0.30\textwidth}
        \centering
        \includegraphics[width=\textwidth]{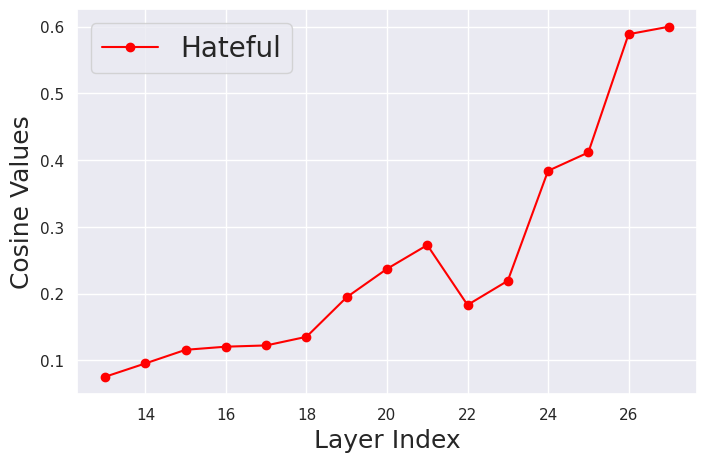}
        \caption{\lama{}: HX}
    \end{subfigure}
    \hfill
    \begin{subfigure}[b]{0.30\textwidth}
        \centering
        \includegraphics[width=\textwidth]{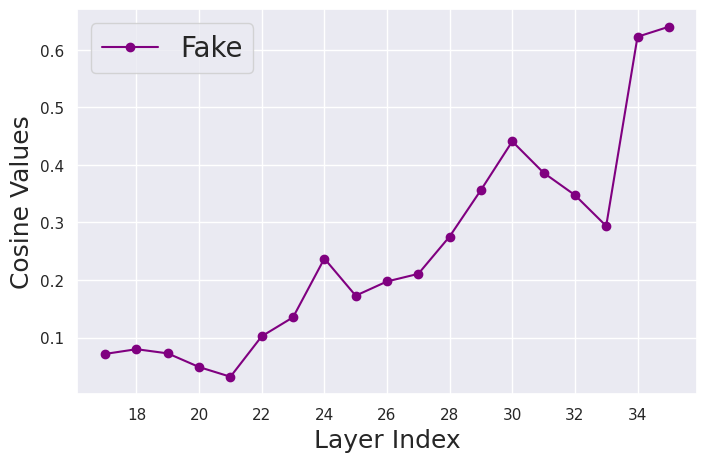}
        \caption{\qwn{}: FE}
    \end{subfigure}

    \vspace{1em} 

    \begin{subfigure}[b]{0.30\textwidth}
        \centering
        \includegraphics[width=\textwidth]{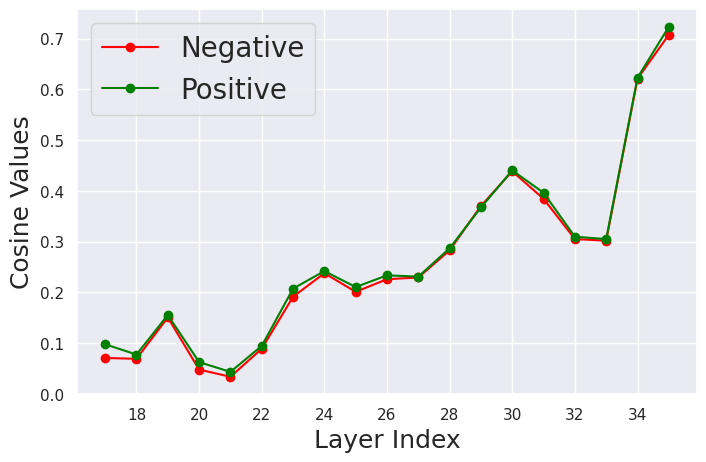}
        \caption{\qwn{}: SNT}
    \end{subfigure}
    \hfill
    \begin{subfigure}[b]{0.30\textwidth}
        \centering
        \includegraphics[width=\textwidth]{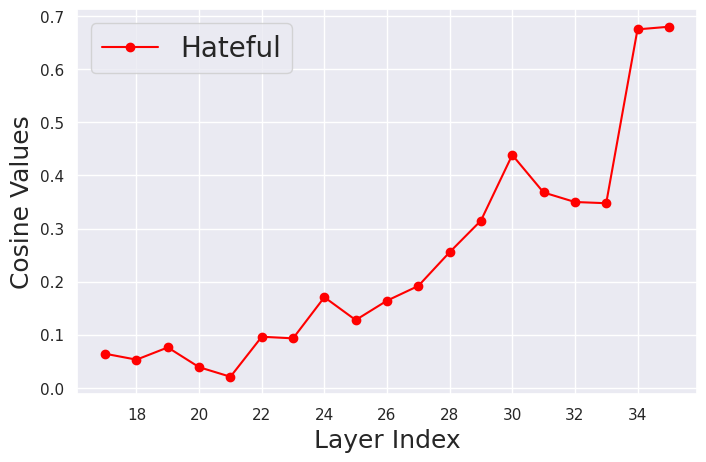}
        \caption{\qwn{}: HX}
    \end{subfigure}
    \hfill
    \begin{subfigure}[b]{0.30\textwidth}
        \centering
        \includegraphics[width=\textwidth]{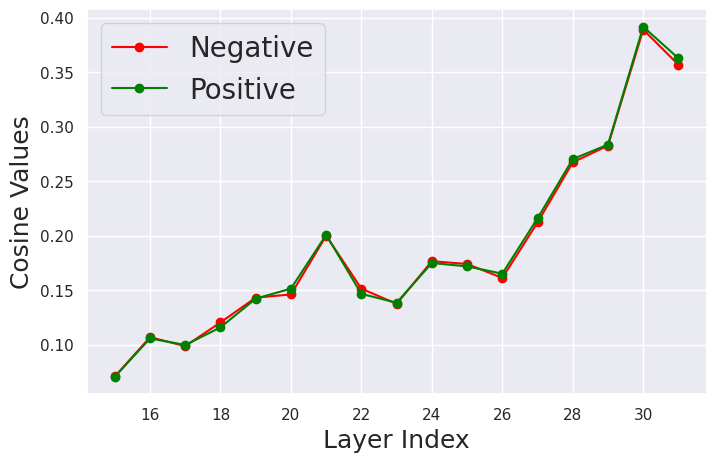}
        \caption{\mist{}: SNT}
    \end{subfigure}

    \caption{\footnotesize Cosine similarity curves $\mathcal{C}$ across different datasets and models. $x$-axis shows the layer indices starting from the middle layer of the model till the end. $y$-axis shows the average cosine similarity of the bias vector with the input token sentences. FE: \fk{}, HX: \hx{}, SNT: \snt{}.}
    \label{cosine_curves}
\end{figure*}

\subsection{Layer selection}
\label{sec:layer_selection}
\begin{algorithm}[!t]
\caption{Heuristic layer selection}
\label{alg:layer_selection}
\begin{algorithmic}[1]
\footnotesize
\Require Fine-tuned LLM $\mathcal{F}$; input sequence $\mathbf{w} = (w_1, w_2, \ldots, w_n)$;
         instruction bias vector $\bm{b}$ (obtained by passing an empty prompt through $\mathcal{F}$);
         cosine threshold $\tau = 0.25$

\Ensure  Selected layer index $\ell^*$

\State Let $L_{\mathrm{mid}}$ denote the middle-indexed decoder layer of $\mathcal{F}$, and
       $L_{\mathrm{last}}$ the final decoder layer.

\For{each layer $\ell \in \{L_{\mathrm{mid}},\, L_{\mathrm{mid}}+1,\, \ldots,\, L_{\mathrm{last}}\}$}
    \State Extract the \texttt{down\_proj} hidden representation $h_t^{(\ell)}$ for each
           token $w_t$, without debiasing.
    \State Compute the per-token cosine similarity with the bias vector:
           \[
               c_t^{(\ell)} \;=\;
               \frac{\langle h_t^{(\ell)},\, \bm{b} \rangle}
                    {\| h_t^{(\ell)} \|\,\| \bm{b} \|}
           \]
    \State Average over all tokens to obtain a layer-level similarity score:
           \[
               \bar{c}^{(\ell)} \;=\; \frac{1}{n}\sum_{t=1}^{n} c_t^{(\ell)}
           \]
\EndFor

\State Form the cosine-similarity curve $\mathcal{C} = \bigl(\bar{c}^{(\ell)}\bigr)_{\ell = L_{\mathrm{mid}}}^{L_{\mathrm{last}}}$.

\State Restrict the candidate set to layers whose similarity lies below the threshold:
       \[
           \mathcal{L}_{\tau} \;=\;
           \bigl\{\, \ell \;\mid\; \bar{c}^{(\ell)} < \tau \,\bigr\}
       \]

\State Within $\mathcal{L}_{\tau}$, compute the discrete gradient of $\mathcal{C}$:
       \[
           \nabla\bar{c}^{(\ell)} \;=\; \bar{c}^{(\ell+1)} - \bar{c}^{(\ell)}
       \]

\State Identify the layer at the lower end of the steepest ascent in $\mathcal{C}$:
       \[
           \ell^* \;=\; \operatorname*{arg\,max}_{\ell \;\in\; \mathcal{L}_{\tau}}
                        \;\nabla\bar{c}^{(\ell)}
       \]

\State \Return $\ell^*$

\end{algorithmic}
\end{algorithm}

\noindent Recall that a transformer has multiple layers, and each layer has its own representation of the input tokens. We define a heuristic to identify the decoder layer whose \texttt{down\_proj} MLP outputs
best represent the actual input sentence, balancing input-token fidelity against
instruction-following bias. The intuition behind this heuristic is as follows. Layers close to the middle layer of the transformer, $L_{\mathrm{mid}}$ tend to be
nearly orthogonal to $\bm{b}$, indicating that their representations are driven primarily by
the actual input tokens. As $\ell$ increases toward $L_{\mathrm{last}}$ (last transformer layer), the cosine similarity also
rises, reflecting a shift in the model's focus toward instruction-following rather
than input content. The selected layer $\ell^*$ sits at the inflection point of this curve $\mathcal{C}$ -- the
point of highest local gradient still below the cosine similarity threshold $\tau = 0.25$ -- where the model strikes the best
balance between encoding input-token semantics and attending to the instruction prompt.
We verify that this heuristic is consistent across all model families and datasets considered
in this work (see example curves $\mathcal{C}$ in Figure~\ref{cosine_curves} for different datasets and models discussed in Section~\ref{sec:expt}). The procedure based on this heuristic is formalized in Algorithm~\ref{alg:layer_selection}, and the layers selected for different model and dataset settings are detailed in Table \ref{tab:optimal_layers}

\begin{table}[htbp]
\centering
\scriptsize 
\setlength{\tabcolsep}{2pt} 
\begin{tabularx}{\columnwidth}{l *{3}{>{\centering\arraybackslash}X}}
\toprule
\textbf{Model Family} & \textbf{Sentiment} & \textbf{\hx{}} & \textbf{\fk{}} \\
\midrule
\lama{} & 19 & 18 & 18 \\
\qwn{}  & 22 & 23 & 23 \\
\mist{} & 20 & 27 & 22 \\
\bottomrule
\end{tabularx}
\caption{Selected optimal layer indices across different datasets and model families.}
\label{tab:optimal_layers}
\end{table}

\subsection{Input token attribution}
We use PyDMD, \cite{Demo2018}, a robust Python package, which consists of implementations of various \dmd{} algorithms, to obtain the approximate surrogate $\mathcal{F}$ of the LLM using the debiased hidden state matrix. The modes obtained after the eigendecomposition of the linear Koopman operator \textbf{A} can be interpreted as capturing low-dimensional structures underlying the evolution of hidden representations. We interpret these structures to reflect linguistic and semantic regularities, such as syntactic patterns, grammatical dependencies, and features relevant to the predictive behavior of the model.\\ 
Apart from the basic \dmd{} algorithm we also use the higher order dynamic mode decomposition (\hdmd{}) \cite{doi:10.1137/15M1054924}. While \dmd{} focuses on finding a linear operator satisfying the Markovian assumption ($h_{t+1} = \mathbf{A}h_t$), \hdmd{} implements time embedding (see Appendix~\ref{app:hdmd} for more details). In this way, a Hankel matrix is constructed from the original snapshot data to model the system such that each hidden state is treated as a linear combination of the $d$ previous time-steps, where $d$ is the delay parameter. To implement \hdmd{}, we stack $d$ successive snapshots into a higher-dimensional representation, which allows the model to capture long-term dynamics and temporal dependencies that standard \dmd{} might miss. The expanded snapshot matrix $\mathcal{H}$ looks as follows

\begin{equation}
\small
\mathcal{H}_{1}^{N-d} = \begin{bmatrix} 
h_1^{\ell^*} & h_2^{\ell^*} & \dots & h^{\ell^*}_{N-d} \\ 
h_2^{\ell^*} & h_3^{\ell^*} & \dots & h_{N-d+1}^{\ell^*} \\ 
\vdots & \vdots & \ddots & \vdots \\ 
h_d^{\ell^*} & h_{d+1}^{\ell^*} & \dots & h_N^{\ell^*} 
\end{bmatrix}
\label{hankel_matrix}
\end{equation}
Our pipeline then proceeds by applying standard \dmd{} to this augmented matrix. First, we perform an SVD on $\mathcal{H}$ to reduce dimensionality while preserving dominant features. Next, we compute the Koopman eigenvalues and eigenvectors from the reduced-rank operator to identify the system's modes.\\
\noindent \textbf{Mode ranking}: Similar to \pca{}, where components are ranked by the amount of variance explained, \dmd{}/\hdmd{} modes can be ranked according to different measures of dynamical significance. We employ two ranking criteria: (i) the magnitude of the initial modal amplitudes \cite{ROWLEY2009}, and (ii) the average magnitude of the modal amplitudes over the temporal trajectory \cite{CRMECA_2014__342_6-7_410_0} to get the most relevant modes (top-$k$) for our task. In all our experiments, we consider $k=5$ modes, both when experimenting with \dmd{} (and other similar baselines like \pca{}).\\
\noindent \textbf{Token ranking}: To find how each token contributes to those modes, we consider the state change ($\Delta h_t^{\ell^*}$) which the token $w_t$ adds after it has been processed, as a proxy for the hidden state representation of that token, and then calculate its scalar projection magnitude on each of the modes. We then sum the projection magnitude values for each mode for all the top $k$ modes to obtain a projection score for that token.  
\begin{align}
\small
\label{delta_h}
\Delta h_t^{\ell^*} &= h_t^{\ell^*} - h_{t-1}^{\ell^*} \\[6pt]
\alpha_{t,i} &= \frac{\left| \langle \Delta h_t^{\ell^*}, \phi_i \rangle \right|}{\|\phi_i\|}, \quad i = 1, \dots, k \\[6pt]
s_t &= \sum_{i=1}^{k} \alpha_{t,i}
\label{attribution}
\end{align}
where, $\phi_i$ is a selected mode, $\langle \cdot \rangle$ represents the dot product of two vectors, $k$ represents the number of top modes considered, and $s_t$ is the projected score obtained for the particular token index $t$.
Finally, we rank the tokens according to their decreasing order of projection scores to obtain the ranked input attributions for each token.

\section{Experimental setup}\label{sec:expt}

\noindent\textbf{Models}: In this work, we fine-tune three different LLMs, \lama{} \cite{grattafiori2024llama}, \qwn{} \cite{yang2025qwen3}, and \mist{} \cite{jiang2023mistral}. We perform supervised instruction fine-tuning of the model in full precision. The prompt templates used for the fine-tuning are provided in the Appendix. \\
\noindent\textbf{Datasets}: For this work, we use three text classification datasets (see Appendix~\ref{app:setting} for more details) as follows -- (i) \snt{}: sentiment classification \cite{zhang2015character}, (ii) \fk{}: fake news detection \cite{nakamura2020fakeddit}, and (iii) \hx{}: hateful content classification \cite{mathew2021hatexplain}. We use the training split of each of these datasets to perform the supervised fine-tuning of the three chosen models. Finally, we use 1000 different sentences from the test set of each dataset, equally divided in different input size buckets, to evaluate the efficacy of our method on small as well as large input sentences.

\noindent\textbf{Baselines}: We compare our method with \ig{} \cite{sundararajan2017axiomatic}, \shap{} \cite{lundberg2017unified} (more specifically, the \textsc{GradientSHAP} implementation) and \pca{}. Note that \pca{} is computed directly on the transpose of debiased hidden state matrix, $\hat{\bm{X}}^\top$, such that each row is treated as an observation (token hidden states) and each column represents the hidden dimension, to obtain a set of principal components that maximizes the variance in the collection of representation vectors. The token attribution in \pca{} then follows the same procedure as \dmd{}, mentioned in Eq. \ref{delta_h} to \ref{attribution}. 
\begin{table*}[t]
\centering
\scriptsize
\begin{tabularx}{\textwidth}{l|X|X}
\toprule
\textbf{Dataset classes} & \textbf{Input sentence} & \textbf{Ground truth (GT)} \\ \midrule
Negative sentiment & hated it it was so small not as the picture showed could not even get it to turn on i threw it in the trash it wasnt worth the hassle of returning it & 1. hated, 2. trash, 3. hassle, 4. not, 5. small, 6. threw, 7. worth, 8. returning \\ 
Positive sentiment & great gift but expensive i got this gift from my husband it is a little expensive but the best gps so far & 1. best, 2. great, 3. gift, 4. husband, 5. gps \\ \midrule
\hx{} (hateful) & user it a refugees fault for sucking up all the oxygen thus my issues get ignored or minimized citizens before regugees & 1. refugees, 2. fault, 3. sucking, 4. oxygen, 5. citizens, 6. regugees \\ \midrule
\fk{} (fake) & womens magazine staff out of work as kardashians reveal they have nothing left to reveal & 1. kardashians, 2. reveal, 3. nothing, 4. left, 5. womens, 6. magazine, 7. staff, 8. out, 9. work \\ \bottomrule
\end{tabularx}
\caption{\footnotesize Examples of input sentences and their corresponding token-level ground truth (GT) rankings generated by GPT-4.1.}
\label{gt_examples}
\end{table*}

\textbf{Metrics}: We report three metrics: (i) average matched count, (ii) Rank-Biased Overlap (RBO) \cite{10.1145/1852102.1852106} and (iii) recall@$k$. The average matched count gives the expected number of tokens matched from the GT, it does not take into account the number of tokens GT has for a particular sentence. RBO compares the rank of the retrieved token with the token's rank in the GT and penalizes tokens that are assigned a higher rank than in the GT. RBO is defined as follows,
\begin{equation}
\small
    \textbf{RBO}_{\text{EXT}}(S, T, p, k)
    =
    (1 - p)\sum_{d=1}^{k} A_d \cdot p^{d-1}
    +
    \frac{X_k}{k} \cdot p^k
\label{rbo}
\end{equation}
where $S$ and $T$ are two retrieved lists, $p$ is the persistence factor, $A_d$ is the agreement between two lists at depth $d$, $k$ is the top-$k$ cut-off value, and $X_k$ is the overlap count between $S$ and $T$ at depth $k$. For our evaluations, we choose $p=0.95$. Recall@$k$ measures the percentage of GT tokens retrieved. It is sensitive to both the number of retrieved tokens and the number of tokens in the GT.\\
\textbf{Ground truth}: For each dataset, ground truth tokens responsible for a prediction were obtained using OpenAI GPT-4.1 \cite{achiam2023gpt}, where the model was prompted to rank individual tokens from input sentences based on their importance to the output label. A maximum of 20 tokens were ranked for each input, apart from stop words, and the model was instructed to stop the ranking early if there were fewer than 20 influential tokens. Table \ref{gt_examples} lists an example sentence from each dataset's class label, and the corresponding ground truth (GT) returned by GPT-4.1. The GTs are ranked in order of the most influential to the least influential in determining the class label of the input sentence. In order to test the goodness of the labelling, two experienced annotators manually rank 100 input sentences from each dataset which is then compared with the GPT-4.1 annotated rankings. We first observe that the two experienced annotators agree very well in terms of the RBO scores for all the three datasets. The scores are as follows -- negative sentiment: 0.64, positive sentiment: 0.72, hateful text: 0.62, and fake text: 0.61. Next, we find that the human annotations and GPT-4.1 annotations also align well. In particular, RBO values obtained for the three datasets are negative sentiment: 0.60, positive sentiment: 0.70, hateful text: 0.54, and fake text: 0.60.   
 The slightly lower agreement for \hx{} corresponds to the known subjectivity of hate speech annotation in the literature.

\begin{table*}[t]
\centering
\scriptsize
\begin{tabularx}{\textwidth}{l|*{3}{C|}|*{3}{C|}}
\toprule
\multirow{2}{*}{\textbf{Method}} & \multicolumn{3}{c|}{\textbf{Negative sentiment}} & \multicolumn{3}{c}{\textbf{Positive sentiment}} \\ 
\cmidrule(lr){2-4} \cmidrule(lr){5-7}
& \textbf{MC@20 $\uparrow$} & \textbf{RBO@20 $\uparrow$} & \textbf{Recall@20 $\uparrow$} & \textbf{MC@20 $\uparrow$} & \textbf{RBO@20 $\uparrow$} & \textbf{Recall@20 $\uparrow$} \\ 
\midrule
\multicolumn{7}{c}{\lama{}} \\ 
\midrule
\ig{} & 4.42 & \cellcolor{lg}\underline{0.24} & 0.60 & 4.69 & \cellcolor{lg}\underline{0.26} & 0.66 \\
\shap{} & 3.85 & 0.20 & 0.47 & 4.26 & 0.21 & 0.50 \\
\pca{} & \cellcolor{lg}\underline{5.24} & \cellcolor{lg}\underline{0.24} & \cellcolor{lg}\underline{0.66} & \cellcolor{lg}\underline{5.98} & \cellcolor{lg}\underline{0.26} & \cellcolor{lg}\underline{0.73} \\
\method{}$^{\hdmd{}-avgamp}$ & \cellcolor{fg}\textbf{5.35} & \cellcolor{fg}\textbf{0.25} & \cellcolor{fg}\textbf{0.68} & \cellcolor{fg}\textbf{6.12} & \cellcolor{fg}\textbf{0.27} & \cellcolor{fg}\textbf{0.76} \\ 
\midrule

\multicolumn{7}{c}{\qwn{}} \\ 
\midrule
\ig{} & 4.60 & \cellcolor{lg}\underline{0.25} & \cellcolor{lg}\underline{0.57} & 5.24 & \cellcolor{lg}\underline{0.28} & 0.63 \\
\shap{} & 3.94 & 0.20 & 0.48 & 4.26 & 0.21 & 0.50 \\
\pca{} & \cellcolor{fg}\textbf{5.46} & \cellcolor{lg}\underline{0.25} & \cellcolor{fg}\textbf{0.69} & \cellcolor{lg}\underline{6.07} & \cellcolor{lg}\underline{0.28} & \cellcolor{lg}\underline{0.74} \\
\method{}$^{\hdmd{}-avgamp}$ & \cellcolor{lg}\underline{5.43} & \cellcolor{fg}\textbf{0.26} & \cellcolor{fg}\textbf{0.69} & \cellcolor{fg}\textbf{6.17} & \cellcolor{fg}\textbf{0.29} & \cellcolor{fg}\textbf{0.76} \\
\midrule

\multicolumn{7}{c}{\mist{}} \\ 
\midrule
\ig{} & 5.03 & \cellcolor{fg}\textbf{0.27} & 0.62 & 5.14 & \cellcolor{lg}\underline{0.26} & 0.61 \\
\shap{} & 4.02 & 0.20 & 0.48 & 4.38 & 0.22 & 0.51 \\
\pca{} & \cellcolor{lg}\underline{5.23} & 0.24 & \cellcolor{lg}\underline{0.65} & \cellcolor{lg}\underline{5.72} & 0.25 & \cellcolor{lg}\underline{0.70} \\
\method{}$^{\hdmd{}-avgamp}$ & \cellcolor{fg}\textbf{5.40} & \cellcolor{lg}\underline{0.25} & \cellcolor{fg}\textbf{0.68} & \cellcolor{fg}\textbf{6.03} & \cellcolor{fg}\textbf{0.27} & \cellcolor{fg}\textbf{0.75} \\
\bottomrule
\end{tabularx}
\caption{\footnotesize Experimental results for the \snt{} dataset across three different model families. MC: Matched count, \hdmd{}-avgamp: \hdmd{} with averaged amplitude ranking. Best results are in \textbf{bold} and the second best are \underline{underlined}.}
\label{sentiment_results}
\end{table*}

\begin{table*}[t]
\centering
\scriptsize
\begin{tabularx}{\textwidth}{l|*{3}{C|}|*{3}{C|}}
\toprule
\multirow{2}{*}{\textbf{Method}} & \multicolumn{3}{c|}{\textbf{Hateful reviews (\hx{})}} & \multicolumn{3}{c}{\textbf{Fake reviews (\fk{})}} \\ 
\cmidrule(lr){2-4} \cmidrule(lr){5-7}
& \textbf{MC@10 $\uparrow$} & \textbf{RBO@10 $\uparrow$} & \textbf{Recall@10 $\uparrow$} & \textbf{MC@10 $\uparrow$} & \textbf{RBO@10 $\uparrow$} & \textbf{Recall@10 $\uparrow$} \\ 
\midrule
\multicolumn{7}{c}{\lama{}} \\ 
\midrule
\ig{} & 2.57 & \cellcolor{lg}\underline{0.24} & \cellcolor{lg}\underline{0.60} & 4.39 & 0.38 & 0.60 \\
\shap{} & 2.30 & 0.22 & 0.50 & 3.60 & 0.32 & 0.48 \\
\pca{} & \cellcolor{lg}\underline{2.63} & 0.23 & \cellcolor{lg}\underline{0.60} & \cellcolor{lg}\underline{5.12} & \cellcolor{fg}\textbf{0.43} & \cellcolor{lg}\underline{0.70} \\
\method{}$^{\dmd{}-amp}$ & \cellcolor{fg}\textbf{2.77} & \cellcolor{fg}\textbf{0.25} & \cellcolor{fg}\textbf{0.64} & \cellcolor{fg}\textbf{5.46} & \cellcolor{lg}\underline{0.42} & \cellcolor{fg}\textbf{0.76} \\ 
\midrule

\multicolumn{7}{c}{\qwn{}} \\ 
\midrule
\ig{} & \cellcolor{lg}\underline{2.66} & \cellcolor{fg}\textbf{0.26} & \cellcolor{lg}\underline{0.60} & 4.35 & 0.38 & 0.60 \\
\shap{} & 2.31 & 0.22 & 0.50 & 3.85 & 0.34 & 0.52 \\
\pca{} & 2.63 &  0.24 & \cellcolor{lg}\underline{0.60} & \cellcolor{lg}\underline{4.90} & \cellcolor{lg}\underline{0.41} & \cellcolor{lg}\underline{0.68} \\
\method{}$^{\dmd{}-amp}$ & \cellcolor{fg}\textbf{2.85} & \cellcolor{lg}\underline{0.25} & \cellcolor{fg}\textbf{0.64} & \cellcolor{fg}\textbf{5.37} & \cellcolor{fg}\textbf{0.42} & \cellcolor{fg}\textbf{0.75} \\
\midrule

\multicolumn{7}{c}{\mist{}} \\ 
\midrule
\ig{} & 2.73 & \cellcolor{lg}\underline{0.26} & 0.62 & 4.43 & 0.39 & 0.61 \\
\shap{} & 2.60 & 0.25 & 0.59 & 4.00 & 0.35 & 0.54 \\
\pca{} & \cellcolor{fg}\textbf{2.95} & \cellcolor{fg}\textbf{0.30} & \cellcolor{fg}\textbf{0.67} & \cellcolor{lg}\underline{5.21} & \cellcolor{fg}\textbf{0.44} & \cellcolor{lg}\underline{0.72} \\
\method{}$^{\dmd{}-amp}$ & \cellcolor{lg}\underline{2.82} & \cellcolor{lg}\underline{0.26} & \cellcolor{lg}\underline{0.65} & \cellcolor{fg}\textbf{5.27} & \cellcolor{lg}\underline{0.41} & \cellcolor{fg}\textbf{0.74} \\
\bottomrule
\end{tabularx}
\caption{\footnotesize Experimental results for the \hx{} and \fk{} datasets across three different model families. The attributions are computed only for the class of interest (i.e., hateful for \hx{} and fake for \fk{}). MC: Matched count, \dmd{}-amp: \dmd{} with amplitude ranking. Best results are in \textbf{bold} and the second best are \underline{underlined}.}
\label{hatexplain_fakeddit_results}
\end{table*}

\section{Results}

We evaluate our framework using 1,000 samples per dataset, partitioned into distinct buckets based on input length to assess performance across varying context sizes. For the \snt{} dataset, samples are categorized into three ranges: 15–40, 40–70, and 70–100 tokens. For the \hx{} and \fk{} datasets, we utilize two buckets: 15–40 and 40–70 tokens.\\ 
As noted earlier, we evaluate four combinations as follows -- (i) standard \dmd{} with amplitude-based ranking, (ii) standard \dmd{} with time-averaged amplitude ranking, (iii) \hdmd{} with amplitude-based ranking using an adaptive delay parameter, and (iv) \hdmd{} with time-averaged amplitude ranking using an adaptive delay parameter. In each setting (model + dataset), we report the metrics obtained for the best performing combination above alongside the baselines (see Appendix~\ref{app:setting} for full results).  
\subsection{Dataset based results} 
\textbf{Sentiment classification}: For sentiment classification, we retrieve the top 20 tokens for both sentiments and then compare them with the GT. We find that \hdmd{} (see Table \ref{sentiment_results}), paired with mode ranking through the averaged amplitude, consistently outperforms all other configurations across the three evaluated models. This setup generally exceeds traditional baselines, including \pca{}, \ig{}, and \shap{} in most metrics. Since sentiment is typically expressed through multi-word phrases rather than isolated tokens, the use of delay factors in the \hdmd{} setup  provides the necessary context to capture these dynamics more effectively than the standard \dmd{}.\\
\noindent \hx{}: For this dataset, the standard \dmd{} using amplitude-based ranking proves to be the most effective configuration for all models. As each data point in \hx{} contains very few sentences with more than 10 GT tokens, we calculate the metrics for the top 10 retrieved tokens. Note that here we compute the attributions on for the class of interest (i.e., hateful class). For \lama{} and \qwn{}, the \dmd{}-based attribution generally outperforms the baseline methods (Table \ref{hatexplain_fakeddit_results}). For \mist{}, our method is the second best. Since \hx{} relies on the specific token-level annotations for hateful content, we believe that the state changes triggered by these individual tokens are well-captured by the standard \dmd{}. This suggests that the evolution of hidden representations in this context follows a near-Markovian process that does not require the extended memory of \hdmd{}. \\
\noindent \fk{}: The results for \fk{} align closely with those of \hx{} (Table \ref{hatexplain_fakeddit_results}). Because the dataset consists of Reddit posts where specific keywords—such as the names of political figures—often dictate the ``fake'' status of a post, the importance is concentrated on individual tokens. Here, again, the attributions are computed for the fake class. Consequently, the standard \dmd{} with amplitude ranking achieves the best performance, outperforming all baseline methods in recall@10. The importance ranking of the tokens may not align best with the GT, but our method retrieves the most amount of tokens from GT as measured by recall. This confirms that for tasks where localized token information is critical, standard linear approximations are highly effective.

\subsection{Qualitative results}

\begin{table*}[t]
\centering
\scriptsize
\renewcommand{\arraystretch}{1.5}
\begin{tabularx}{\textwidth}{l|l|X}
\toprule
\textbf{Model} & \textbf{Method} & \textbf{Top-ranked Tokens (Attribution)} \\
\midrule
\multicolumn{3}{p{\textwidth}}{``dont buy this phone this phone makes a constant high static shrieking noise i bought this phone to use on a regularly scheduled long conference call and was totally unable to use it its going directly in the trash" \textbf{[Negative]}} \\
\midrule
\multirow{4}{*}{\lama{}}
 & GT         & 1. trash, 2. unable, 3. shrieking, 4. static, 5. noise, 6. constant, 7. dont, 8. buy, 9. totally, 10. directly \\
 & \method{} & 1. going, 2. trash, 3. dont, 4. unable, 5. conference, 6. shrieking, 7. makes, 8. bought, 9. regularly, 10. high, 11. scheduled, 12. use, 13. long, 14. directly, 15. phone, 16. constant, 17. call, 18. static, 19. totally \\
 & \pca{}        & 1. going, 2. unable, 3. conference, 4. shrieking, 5. makes, 6. high, 7. bought, 8. trash, 9. phone, 10. regularly, 11. constant, 12. static, 13. long, 14. use, 15. noise, 16. directly \\
 & \ig{}         & 1. scheduled, 2. dont, 3. conference, 4. use, 5. totally, 6. trash, 7. directly, 8. going, 9. use, 10. static, 11. call, 12. unable, 13. noise, 14. buy, 15. regularly, 16. long \\
 & \shap{}       & 1. trash, 2. directly, 3. noise, 4. high, 5. scheduled, 6. use, 7. constant, 8. going, 9. static, 10. dont, 11. phone, 12. totally \\
\midrule
\multirow{4}{*}{\qwn{}}
 & GT         & 1. trash, 2. unable, 3. shrieking, 4. static, 5. noise, 6. constant, 7. dont, 8. buy, 9. totally, 10. directly \\
 & \method{} & 1. going, 2. directly, 3. trash, 4. dont, 5. regularly, 6. makes, 7. buy, 8. static, 9. unable, 10. bought, 11. scheduled, 12. use, 13. totally, 14. long, 15. phone, 16. shrieking, 17. conference, 18. constant, 19. noise \\
 & \pca{}        & 1. dont, 2. buy, 3. going, 4. bought, 5. makes, 6. regularly, 7. shrieking, 8. trash, 9. constant, 10. scheduled, 11. static, 12. noise, 13. directly, 14. use, 15. conference, 16. totally, 17. call, 18. long, 19. unable \\
 & \ig{}         & 1. use, 2. dont, 3. unable, 4. regularly, 5. totally, 6. conference, 7. scheduled, 8. use, 9. noise, 10. directly, 11. going, 12. call, 13. constant, 14. phone, 15. bought, 16. high, 17. static, 18. phone, 19. trash \\
 & \shap{}       & 1. use, 2. high, 3. conference, 4. totally, 5. dont, 6. regularly, 7. makes, 8. phone, 9. buy, 10. scheduled, 11. directly, 12. shrieking \\
\midrule
\multicolumn{3}{p{\textwidth}}{``you are retarded and do not know what reply mean or are a shitty troller suck a bag of ni**er d*cks fa**ot" \textbf{[Hateful]}} \\
\midrule
\multirow{4}{*}{\lama{}}
 & GT         & 1. ni**er, 2. fa**ot, 3. retarded, 4. shitty, 5. d*cks \\
 & \method{} & 1. fa**ot, 2. d*cks, 3. suck, 4. ni**er, 5. bag, 6. troller, 7. shitty, 8. retarded, 9. mean \\
 & \pca{}        & 1. suck, 2. reply, 3. know, 4. retarded, 5. troller, 6. bag, 7. d*cks, 8. shitty, 9. mean, 10. fa**ot, 11. ni**er \\
 & \ig{}        & 1. shitty, 2. fa**ot, 3. troller, 4. mean, 5. know, 6. bag, 7. ni**er \\
 & \shap{}      & 1. fa**ot, 2. retarded, 3. know, 4. ni**ger, 5. bag, 6. shitty \\
\midrule
\multirow{4}{*}{\qwn{}}
 & GT         & 1. ni**er, 2. fa**ot, 3. retarded, 4. shitty, 5. d*cks \\
 & \method{} & 1. fa**ot, 2. d*cks, 3. bag, 4. suck, 5. ni**er, 6. troller, 7. reply, 8. mean, 9. shitty, 10. know, 11. retarded \\
 & \pca{}       & 1. retarded, 2. d*cks, 3. fa**ot, 4. bag, 5. suck, 6. know, 7. reply, 8. mean, 9. troller, 10. shitty, 11. ni**er \\
 & \ig{}         & 1. bag, 2. ni**er, 3. retarded, 4. fa**ot, 5. suck, 6. troller, 7. know, 8. d*cks, 9. shitty \\
 & \shap{}       & 1. fa**ot, 2. retarded, 3. reply, 4. ni**er, 5. d*cks\\
\bottomrule
\end{tabularx}
\caption{\footnotesize Attribution for \lama{} and \qwn{} models on negative sentiment and hate speech sentences.}
\label{tab:qualitative_comparison_merged}
\end{table*}

Some of the representative qualitative results are noted in Table~\ref{tab:qualitative_comparison_merged}. \method{} consistently demonstrates the strongest alignment with GT attributions across both models and sentence types, ranking the most semantically relevant tokens near the top of its lists. For the negative sentence, \method{} places ``trash'' and ``unable'' within its top \textit{four} for \lama{}, while for the hate speech it recovers all five GT tokens within its top \textit{nine} for both models. Notably, \method{} also exhibits strong cross-model consistency -- its top five tokens for hate speech are nearly identical between \lama{} and \qwn{}, differing only in minor reordering. \shap{} performs competitively, recovering several high-priority GT tokens in compact lists: for the hate speech sentence it ranks ``fa**ot'' and ``retarded'' in its top two for both models, and for the negative sentence it places ``trash'' first for \lama{}. However, \shap{} tends to surface contextually plausible but GT-absent tokens (e.g. ``high'', ``conference'', ``scheduled'') at the expense of key sentiment markers such as ``unable'' and ``shrieking''. \pca{} shows moderate but uneven alignment, capturing several GT tokens yet struggling to prioritize the most discriminative ones (e.g., ranking ``ni**er'', the highest GT token, last in the hate speech sentence for \lama{}). \ig{} shows the weakest performance overall, with non-salient tokens frequently appearing at the top of its rankings and instability manifesting as duplicate token entries (e.g., ``use'' appears twice in its \qwn{} negative sentence list).\\
Beyond ranking quality, the methods also differ in coverage and consistency. \method{} and \pca{} produce longer attribution lists, surfacing a broader set of contributing tokens, while \shap{} and \ig{} yield shorter and more concentrated lists. This compactness is a strength for \shap{}, which tends to maintain reasonable precision within its smaller set, but a liability for \ig{}, which risks omitting relevant tokens such as ``noise'' and ``constant'' in the negative sentence. Across models, \pca{} and \ig{} exhibit greater sensitivity to the underlying model's representations --- for instance, \pca{}'s top-ranked tokens for the negative sentence differ substantially between \lama{} and \qwn{} --- whereas \method{} and \shap{} remain comparatively stable. Taken together, these observations suggest that \method{} offers the best balance of GT recall, coverage, and cross-model robustness, with \shap{} being a reasonable but less comprehensive alternative, and \pca{} and \ig{} largely lagging behind.

\subsection{Sensitivity to layer selection} 

Here we investigate how sensitive \method{} is to the layer selected (see Algorithm~\ref{alg:layer_selection}) for obtaining our results. We report the recall values for each of the datasets and models in Table~\ref{tab:layer_sens_all}. We note that \method{} is quite stable, and the results using $\ell^*-1$ or $\ell^*+1$ are very similar to those using $\ell^*$. In fact, as long as the cosine similarity of the bias and the selected layer is close to that between the bias and $\ell^*$, the results remain largely unchanged.

\begin{table*}[t] 
\centering
\scriptsize
\setlength{\tabcolsep}{2.5pt}

\begin{tabularx}{\textwidth}{l | *{2}{C} | *{2}{C} || *{2}{C}}
\toprule
\multirow{2}{*}{\textbf{Method}} & \multicolumn{4}{c||}{\textbf{Sentiment Analysis}} & \multicolumn{2}{c}{\textbf{HateXplain}} \\ 
\cmidrule(lr){2-5} \cmidrule(lr){6-7}
& \multicolumn{2}{c|}{\textbf{Negative Sentiment}} & \multicolumn{2}{c||}{\textbf{Positive Sentiment}} & \multicolumn{2}{c}{\textbf{Hateful Label}} \\ 
\cmidrule(lr){2-3} \cmidrule(lr){4-5} \cmidrule(lr){6-7}
& \textbf{Acc Drop (\%) $\uparrow$} & \textbf{Conf Drop $\uparrow$} & \textbf{Acc Drop (\%) $\uparrow$} & \textbf{Conf Drop $\uparrow$} & \textbf{Acc Drop (\%) $\uparrow$} & \textbf{Conf Drop $\uparrow$} \\ 
\midrule

\multicolumn{7}{c}{\textbf{\lama{}}} \\ 
\midrule
IG & 12.18 & 0.15 & 15.17 & 0.20 & 50.66 & 0.34 \\
PCA & \cellcolor{fg}\textbf{13.68} & \cellcolor{lg}\underline{0.17} & \cellcolor{lg}\underline{18.67} & \cellcolor{lg}\underline{0.24} & \cellcolor{fg}\textbf{56.88} & \cellcolor{fg}\textbf{0.37} \\
\method{} & \cellcolor{lg}\underline{12.57} & \cellcolor{fg}\textbf{0.18} & \cellcolor{fg}\textbf{19.26} & \cellcolor{fg}\textbf{0.25} & \cellcolor{lg}\underline{45.50} & \cellcolor{lg}\underline{0.32} \\ 
\midrule

\multicolumn{7}{c}{\textbf{\qwn{}}} \\ 
\midrule
IG & \cellcolor{fg}\textbf{15.77} & \cellcolor{fg}\textbf{0.19} & 19.07 & 0.24 & \cellcolor{fg}\textbf{57.01} & \cellcolor{fg}\textbf{0.37} \\
PCA & \cellcolor{lg}\underline{12.68} & \cellcolor{fg}\textbf{0.19} & \cellcolor{fg}\textbf{26.15} & \cellcolor{fg}\textbf{0.28} & \cellcolor{lg}\underline{47.75} & \cellcolor{lg}\underline{0.32} \\
\method{} & 10.78 & \cellcolor{lg}\underline{0.18} & \cellcolor{lg}\underline{25.35} & \cellcolor{fg}\textbf{0.28} & 44.31 & 0.31 \\ 
\midrule

\multicolumn{7}{c}{\textbf{\mist{}}} \\ 
\midrule
IG & \cellcolor{fg}\textbf{26.75} & \cellcolor{fg}\textbf{0.27} & 8.48 & 0.08 & 34.79 & 0.22 \\
PCA & \cellcolor{lg}\underline{19.76} & \cellcolor{lg}\underline{0.20} & \cellcolor{lg}\underline{20.26} & \cellcolor{lg}\underline{0.19} & \cellcolor{fg}\textbf{51.19} & \cellcolor{fg}\textbf{0.33} \\
\method{} & 17.76 & 0.19 & \cellcolor{fg}\textbf{21.66} & \cellcolor{fg}\textbf{0.21} & \cellcolor{lg}\underline{44.53} & \cellcolor{lg}\underline{0.29} \\
\bottomrule
\end{tabularx}
\caption{Comparison of accuracy drop (\%) and confidence score drop under $top\mathchar`-k$ token masking across the models on \snt{} and \hx{} datasets. Higher drops ($\uparrow$) indicate more faithful attributions.}
\label{tab:performance_drops}
\end{table*}

\begin{table}[ht]
\centering
\scriptsize
\setlength{\tabcolsep}{2pt}
\begin{tabularx}{\columnwidth}{l|l|*{3}{C}|C}
\toprule
\textbf{Model} & \textbf{Dataset} & $\ell^*-1$ & $\ell^*$ & $\ell^*+1$ & \textbf{$\Delta_{\text{max}}$} \\
\midrule
\multirow{4}{*}{\lama{}}
 & Negative & 0.69 & 0.68 & 0.68 & 0.01 \\
 & Positive & 0.76 & 0.75 & 0.75 & 0.01 \\
 & Hateful  & 0.64 & 0.64 & 0.63 & 0.01 \\
 & Fake     & 0.76 & 0.76 & 0.76 & 0.00 \\
\midrule
\multirow{4}{*}{\qwn{}}
 & Negative & 0.68 & 0.69 & 0.68 & 0.01 \\
 & Positive & 0.74 & 0.76 & 0.75 & 0.02 \\
 & Hateful  & 0.62 & 0.64 & 0.62 & 0.02 \\
 & Fake     & 0.75 & 0.75 & 0.75 & 0.00 \\
\midrule
\multirow{4}{*}{\mist}
 & Negative & 0.67 & 0.67 & 0.67 & 0.00 \\
 & Positive & 0.74 & 0.74 & 0.73 & 0.01 \\
 & Hateful  & 0.65 & 0.65 & 0.64 & 0.01 \\
 & Fake     & 0.73 & 0.74 & 0.74 & 0.01 \\
\bottomrule
\end{tabularx}
\caption{\footnotesize Layer-wise sensitivity analysis. $\Delta_{\text{max}}$ represents the maximum DMD performance difference observed from layer $\ell^*$ to either $\ell^*-1$ or $\ell^*+1$.}
\label{tab:layer_sens_all}
\end{table}

\subsection{Fidelity based analysis}\label{fidelity}
To evaluate the faithfulness of our input attribution framework, we conduct a token-masking experiment following established perturbation benchmarks. Specifically, we measure the drop in model accuracy and output confidence when masking the top-$k$ most influential tokens as identified by each attribution method, with whitespace characters. As reported in Table \ref{tab:performance_drops}, our approach consistently ranks first or second in both accuracy and confidence drops across all evaluated model families and datasets, confirming its robust capability to locate the most critical input features. Notably, while gradient-based baselines like \ig{} achieve high accuracy drops on \hx{} by isolating explicit target terms, they often focus strictly on a narrow subset of toxic keywords. This behavior is corroborated by higher RBO scores alongside lower recall values in Table \ref{hatexplain_fakeddit_results}, indicating that \ig{} overlooks implicit, contextually essential tokens. In contrast, our method captures a broader, more cohesive set of influential features while maintaining competitive perturbation performance.

\section{Conclusion}

In this paper, we proposed a framework for interpreting the predictions of fine-tuned LLMs by analyzing the evolution of their hidden states during sequence processing. By treating the model's internal mechanics as a dynamical system, we demonstrate that \dmd{} provides a more effective lens for input attribution than traditional methods. 
Our findings suggest that this dynamical perspective offers a better and more robust way to understand how LLMs represent and process complex information compared to state-of-the-art methods.   

\section*{Acknowledgment}
The authors gratefully acknowledge the financial and institutional support provided by the Joint PhD Programme between the Indian Institute of Technology Kharagpur (IIT Kharagpur) and The University of Manchester.

\section{Limitations}

In this section, we discuss potential limitations of our approach in this work for interpreting LLMs. First, we only fine-tune and evaluate LLMs for classification tasks, where there are very few tokens (1-2) generated as a response to the input query. While this makes our approach efficient for identifying key tokens from the input when only one to two tokens are generated as output, it is still a challenge to find out the influential tokens from the input if more tokens are generated as these tokens also get fed to the model in an auto-regressive manner to generate the next token. For example, in machine translation from English to French, one can find the tokens that dominate the low dimensional structures for the first few French tokens generated, but as these French tokens are also fed to the model to generate the complete translation, formulating the \dmd{} matrix in this case becomes somewhat difficult. So accurately mapping the English tokens responsible for French translated tokens becomes a challenging task and is potential future work.

\noindent Second, in this work we only explore supervised instruction fine-tuned LLMs for simple downstream tasks, but the general purpose instruction models released by the companies remains unexplored in this study. These instruction tuned models are base models fine-tuned on huge datasets for instruction following tasks, and identifying key patterns in their input processing is also a potential future work.

\bibliography{latex/custom}

@article{vaswani2017attention,
  title={Attention is all you need},
  author={Vaswani, Ashish and Shazeer, Noam and Parmar, Niki and Uszkoreit, Jakob and Jones, Llion and Gomez, Aidan N and Kaiser, {\L}ukasz and Polosukhin, Illia},
  journal={Advances in Neural Information Processing Systems},
  volume={30},
  year={2017}
}

@article{jiang2023mistral,
  title={Mistral 7B},
  author={Jiang, Albert Q and Sablayrolles, Alexandre and Mensch, Arthur and Bamford, Chris and Chaplot, Devendra Singh and Casas, Diego de las and Bressand, Florian and Lengyel, Guillaume and Lample, Guillaume and Saulnier, Lucile and others},
  journal={arXiv:2310.06825},
  year={2023}
}

@inproceedings{devlin2019bert,
  title={Bert: Pre-training of deep bidirectional transformers for language understanding},
  author={Devlin, Jacob and Chang, Ming-Wei and Lee, Kenton and Toutanova, Kristina},
  booktitle={Proceedings of the 2019 conference of the North American chapter of the Association for Computational Linguistics: Human Language Technologies, volume 1 (long and short papers)},
  pages={4171--4186},
  year={2019}
}

@article{radford2018improving,
  title={Improving language understanding by generative pre-training},
  author={Radford, Alec and Narasimhan, Karthik and Salimans, Tim and Sutskever, Ilya and others},
  year={2018},
  publisher={San Francisco, CA, USA}
}

@inproceedings{tenney-etal-2019-bert,
    title = "{BERT} Rediscovers the Classical {NLP} Pipeline",
    author = "Tenney, Ian  and
      Das, Dipanjan  and
      Pavlick, Ellie",
    editor = "Korhonen, Anna  and
      Traum, David  and
      M{\`a}rquez, Llu{\'i}s",
    booktitle = "Proceedings of the 57th Annual Meeting of the Association for Computational Linguistics",
    month = jul,
    year = "2019",
    address = "Florence, Italy",
    publisher = "Association for Computational Linguistics",
    url = "https://aclanthology.org/P19-1452/",
    doi = "10.18653/v1/P19-1452",
    pages = "4593--4601"
}

@article{olsson2022context,
  title={In-context learning and induction heads},
  author={Olsson, Catherine and Elhage, Nelson and Nanda, Neel and Joseph, Nicholas and DasSarma, Nova and Henighan, Tom and Mann, Ben and Askell, Amanda and Bai, Yuntao and Chen, Anna and others},
  journal={arXiv preprint arXiv:2209.11895},
  year={2022}
}

@inproceedings{geva2021transformer,
  title={Transformer feed-forward layers are key-value memories},
  author={Geva, Mor and Schuster, Roei and Berant, Jonathan and Levy, Omer},
  booktitle={Proceedings of the 2021 Conference on Empirical Methods in Natural Language Processing},
  pages={5484--5495},
  year={2021}
}

@article{meng2022locating,
  title={Locating and editing factual associations in gpt},
  author={Meng, Kevin and Bau, David and Andonian, Alex and Belinkov, Yonatan},
  journal={Advances in Neural Information Processing Systems},
  volume={35},
  pages={17359--17372},
  year={2022}
}

@article{schmid2010dynamic,
  title={Dynamic mode decomposition of numerical and experimental data},
  author={Schmid, Peter J},
  journal={Journal of Fluid Mechanics},
  volume={656},
  pages={5--28},
  year={2010},
  publisher={Cambridge University Press}
}

@article{Demo2018,
    author = {Nicola Demo and Marco Tezzele and Gianluigi Rozza},
    title = {{P}y{DMD}: {P}ython Dynamic Mode Decomposition},
    journal = {Journal of Open Source Software},
    year = {2018},
    publisher = {The Open Journal},
    volume = {3},
    number = {22},
    pages = {530},
    doi = {10.21105/joss.00530},
    url = {https://doi.org/10.21105/joss.00530},
}

@article{ROWLEY2009, title={Spectral analysis of nonlinear flows}, volume={641}, DOI={10.1017/S0022112009992059}, journal={Journal of Fluid Mechanics}, author={Rowley, Clarence W. and Mezic, Igor and Bagheri, Shervin and Schlatter, Phillpp and Henningson, Dan S.}, year={2009}, pages={115–127}}

@article{CRMECA_2014__342_6-7_410_0,
     author = {Gilles Tissot and Laurent Cordier and Nicolas Benard and Bernd R. Noack},
     title = {Model reduction using {Dynamic} {Mode} {Decomposition}},
     journal = {Comptes Rendus. M\'ecanique},
     pages = {410--416},
     year = {2014},
     publisher = {Elsevier},
     volume = {342},
     number = {6-7},
     doi = {10.1016/j.crme.2013.12.011},
     language = {en},
}

@article{grattafiori2024llama,
  title={The llama 3 herd of models},
  author={Grattafiori, Aaron and Dubey, Abhimanyu and Jauhri, Abhinav and Pandey, Abhinav and Kadian, Abhishek and Al-Dahle, Ahmad and Letman, Aiesha and Mathur, Akhil and Schelten, Alan and Vaughan, Alex and others},
  journal={arXiv preprint arXiv:2407.21783},
  year={2024}
}

@article{yang2025qwen3,
  title={Qwen3 technical report},
  author={Yang, An and Li, Anfeng and Yang, Baosong and Zhang, Beichen and Hui, Binyuan and Zheng, Bo and Yu, Bowen and Gao, Chang and Huang, Chengen and Lv, Chenxu and others},
  journal={arXiv preprint arXiv:2505.09388},
  year={2025}
}

@article{zhang2015character,
  title={Character-level convolutional networks for text classification},
  author={Zhang, Xiang and Zhao, Junbo and LeCun, Yann},
  journal={Advances in Neural Information Processing Systems},
  volume={28},
  year={2015}
}

@inproceedings{nakamura2020fakeddit,
  title={Fakeddit: A new multimodal benchmark dataset for fine-grained fake news detection},
  author={Nakamura, Kai and Levy, Sharon and Wang, William Yang},
  booktitle={Proceedings of the twelfth language resources and evaluation conference},
  pages={6149--6157},
  year={2020}
}

@inproceedings{mathew2021hatexplain,
  title={Hatexplain: A benchmark dataset for explainable hate speech detection},
  author={Mathew, Binny and Saha, Punyajoy and Yimam, Seid Muhie and Biemann, Chris and Goyal, Pawan and Mukherjee, Animesh},
  booktitle={Proceedings of the AAAI conference on artificial intelligence},
  volume={35},
  number={17},
  pages={14867--14875},
  year={2021}
}

@inproceedings{sundararajan2017axiomatic,
  title={Axiomatic attribution for deep networks},
  author={Sundararajan, Mukund and Taly, Ankur and Yan, Qiqi},
  booktitle={International Conference on Machine Learning},
  pages={3319--3328},
  year={2017},
  organization={PMLR}
}

@article{lundberg2017unified,
  title={A unified approach to interpreting model predictions},
  author={Lundberg, Scott M and Lee, Su-In},
  journal={Advances in Neural Information Processing Systems},
  volume={30},
  year={2017}
}

@article{doi:10.1137/15M1054924,
author = {Le Clainche, Soledad and Vega, Jos\'{e} M.},
title = {Higher Order Dynamic Mode Decomposition},
journal = {SIAM Journal on Applied Dynamical Systems},
volume = {16},
number = {2},
pages = {882-925},
year = {2017},
doi = {10.1137/15M1054924},

URL = { 
        https://doi.org/10.1137/15M1054924

},
eprint = { 
        https://doi.org/10.1137/15M1054924

}
}

@article{hernandez2023linearity,
  title={Linearity of relation decoding in transformer language models},
  author={Hernandez, Evan and Sharma, Arnab Sen and Haklay, Tal and Meng, Kevin and Wattenberg, Martin and Andreas, Jacob and Belinkov, Yonatan and Bau, David},
  journal={arXiv preprint arXiv:2308.09124},
  year={2023}
}

@article{achiam2023gpt,
  title={Gpt-4 technical report},
  author={Achiam, Josh and Adler, Steven and Agarwal, Sandhini and Ahmad, Lama and Akkaya, Ilge and Aleman, Florencia Leoni and Almeida, Diogo and Altenschmidt, Janko and Altman, Sam and Anadkat, Shyamal and others},
  journal={arXiv preprint arXiv:2303.08774},
  year={2023}
}

@article{10.1145/1852102.1852106,
author = {Webber, William and Moffat, Alistair and Zobel, Justin},
title = {A similarity measure for indefinite rankings},
year = {2010},
issue_date = {November 2010},
publisher = {Association for Computing Machinery},
address = {New York, NY, USA},
volume = {28},
number = {4},
issn = {1046-8188},
url = {https://doi.org/10.1145/1852102.1852106},
doi = {10.1145/1852102.1852106},
journal = {ACM Trans. Inf. Syst.},
month = nov,
articleno = {20},
numpages = {38}
}

@article{hemati2017biasing,
  title={De-biasing the dynamic mode decomposition for applied Koopman spectral analysis of noisy datasets},
  author={Hemati, Maziar S and Rowley, Clarence W and Deem, Eric A and Cattafesta, Louis N},
  journal={Theoretical and Computational Fluid Dynamics},
  volume={31},
  number={4},
  pages={349--368},
  year={2017},
  publisher={Springer}
}

@article{kokhlikyan2020captum,
  title={Captum: A unified and generic model interpretability library for pytorch},
  author={Kokhlikyan, Narine and Miglani, Vivek and Martin, Miguel and Wang, Edward and Alsallakh, Bilal and Reynolds, Jonathan and Melnikov, Alexander and Kliushkina, Natalia and Araya, Carlos and Yan, Siqi and others},
  journal={arXiv preprint arXiv:2009.07896},
  year={2020}
}

@article{belrose2023eliciting,
  title={Eliciting latent predictions from transformers with the tuned lens},
  author={Belrose, Nora and Ostrovsky, Igor and McKinney, Lev and Furman, Zach and Smith, Logan and Halawi, Danny and Biderman, Stella and Steinhardt, Jacob},
  journal={arXiv preprint arXiv:2303.08112},
  year={2023}
}

@article{erichson2019compressed,
  title={Compressed dynamic mode decomposition for background modeling},
  author={Erichson, N Benjamin and Brunton, Steven L and Kutz, J Nathan},
  journal={Journal of Real-Time Image Processing},
  volume={16},
  number={5},
  pages={1479--1492},
  year={2019},
  publisher={Springer}
}

@inproceedings{kutz2017dynamic,
  title={Dynamic mode decomposition for background modeling},
  author={Kutz, J Nathan and Erichson, N Benjamin and Askham, Travis and Pendergrass, Seth and Brunton, Steven L},
  booktitle={Proceedings of the 16th IEEE International Conference on Computer Vision (ICCV), Venice, Italy},
  pages={22--29},
  year={2017}
}

@incollection{sachin2019dynamic,
  title={Dynamic mode-based feature with random mapping for sentiment analysis},
  author={Sachin Kumar, S and Anand Kumar, M and Soman, KP and Poornachandran, Prabaharan},
  booktitle={Intelligent Systems, Technologies and Applications: Proceedings of ISTA 2018},
  pages={1--15},
  year={2019},
  publisher={Springer}
}

@article{mao2020eigenemo,
  title={EigenEmo: Spectral utterance representation using dynamic mode decomposition for speech emotion classification},
  author={Mao, Shuiyang and Ching, PC and Lee, Tan},
  journal={arXiv preprint arXiv:2008.06665},
  year={2020}
}

@article{vyshnav2020offensive,
  title={Offensive language detection: A comparative analysis},
  author={Vyshnav, MT and Kumar, Sachin and Soman, KP},
  journal={arXiv preprint arXiv:2001.03131},
  volume={10},
  year={2020}
}

\appendix

\section{Use of AI assistance}
We employed proprietary LLMs solely for obtaining the ground truth labels for the datasets, editorial purposes, including refining grammar, spelling, word choice, and overall clarity of the manuscript.

\section{Dynamic mode decomposition}\label{app:dmd}

Dynamic mode decomposition was introduced in \cite{schmid2010dynamic}, to extract dynamic patterns from a flow field in fluid mechanics. \dmd{} works on observational data, without any underlying assumptions of the dynamical system. 
Consider a set of observations obtained from a dynamical system, represented by the matrix $\bm{X}_1^N$, 
$$\bm{X}_1^N = [x_1, x_2, ..., x_N]$$
where each column vector $\bm{x}_i$ represents an $i^{th}$ observation of size \textit{m}. Next, the assumption is that, there exists a linear mapping \textbf{A} which connects the observation $x_i$ to $x_{i+1}$ as follows
$$x_{i+1} = \textbf{A}x_i$$
In the matrix notation we can write the above equation as,
\begin{equation}
    \bm{X}_2^N \approx \textbf{A}\bm{X}_1^{N-1}
\label{dmd_setup}
\end{equation}
where, $\bm{X}_1^{N-1} \in m \times n-1$, is the matrix containing the first $N-1$ snapshots, and $\bm{X}_2^N$ is the matrix shifted by one time-step. Generally, the spatial dimension of the systems is much larger than the temporal dimension, $m \gg n$ and hence Eq \ref{dmd_setup} can be solved using the singular-value decomposition (SVD): 
 \begin{equation}
     X = U\Sigma V^*
 \end{equation}
where * represents complex conjugate transpose, $U \in \mathbb{C}^{n \times r}$, $\Sigma \in \mathbb{C}^{r \times r}$, and $V \in \mathbb{C}^{r \times n}$.
Hence
\begin{equation}
    \textbf{A} = \bm{X}_2^N V \Sigma^{-1} U^*
\end{equation}
as computing this can be computationally expensive, we project \textbf{A} onto the lower dimensional subspace defined by the eigenvectors of the snapshot matrix, $\tilde{\textbf{A}} = U^*\textbf{A}U$ and then again define the \dmd{} problem as,
\begin{equation}
    \tilde{\textbf{x}}_{t+1} = \tilde{\textbf{A}} \tilde{\textbf{x}}_t
\end{equation}
After computing $\tilde{\textbf{A}}$, the eigenvector-decomposition is done and then the lower dimensional eigenvectors are projected back to the original spatial dimension, to approximate the eigenvectors of \textbf{A}.
\begin{equation}
    \tilde{\textbf{A}} \mathbf{W} = \mathbf{W} \Lambda
\label{eigen_prob}
\end{equation}
\begin{equation}
    \boldsymbol{\Phi} = \bm{X}_2^N V \Sigma^{-1} \mathbf{W}
\end{equation}
The eigenvectors obtained from Eq \ref{eigen_prob}, are known as \dmd{} modes, and they represent the dominant patterns which construct the flow of the dynamical system. Each eigenvector $\phi_i$, is associated with a complex eigenvalue which represents the strength and the frequency of oscillations of that eigenvector.

\section{Higher order dynamic mode decomposition}\label{app:hdmd}

\hdmd{} exploits a delay-embedding approach which helps it capture temporal dependencies across multiple snapshots. Consider a set of observations $\bm{X}_1^N = [x_1, x_2, \dots, x_N]$. The core assumption of \hdmd{} is that there exists a higher-order linear relationship such that the observation $x_{i+d}$ is a linear combination of the previous $d$ snapshots:
\begin{equation}
    x_{i+d} \approx \mathbf{A}_0 x_i + \mathbf{A}_1 x_{i+1} + \dots + \mathbf{A}_{d-1} x_{i+d-1}
\end{equation}
where $d$ is the design parameter representing the number of delays. This is solved by constructing a Hankel snapshot matrix $\mathcal{H}_1^{N-d+1}$ by stacking $d$ successive snapshots into a higher-dimensional representation as follows.
\begin{equation}
\mathcal{H}_1^{N-d+1} = \begin{bmatrix} 
x_1 & x_2 & \dots & x_{N-d+1} \\ 
x_2 & x_3 & \dots & x_{N-d+2} \\ 
\vdots & \vdots & \ddots & \vdots \\ 
x_d & x_{d+1} & \dots & x_N 
\end{bmatrix}
\end{equation}
In this augmented space, the system can be treated as a first-order dynamical system:
\begin{equation}
    \mathcal{H}_2^{N-d+1} \approx \mathbf{\mathcal{A}} \mathcal{H}_1^{N-d}
\end{equation}
where $\mathcal{H}_2^{N-d+1}$ is the Hankel matrix shifted by one time-step. To efficiently compute the operator $\mathbf{\mathcal{A}}$, the SVD is performed on the Hankel matrix similar to \dmd{}.
\begin{equation}
    \mathcal{H}_1^{N-d} = \mathcal{U} \Sigma \mathcal{V}^*
\end{equation}
The high-dimensional operator is then projected onto the lower-dimensional subspace defined by the SVD modes of the Hankel matrix, yielding the reduced-order operator $\tilde{\mathbf{A}}$:
\begin{equation}
    \tilde{\mathbf{A}} = \mathcal{U}^* \mathcal{H}_2^{N-d+1} \mathcal{V} \Sigma^{-1}
\end{equation}
The reduced operator is then eigendecomposed as:
\begin{equation}
    \tilde{\mathbf{A}} \mathbf{W} = \mathbf{W} \Lambda
\end{equation}
The \hdmd{} modes are obtained by projecting these eigenvectors back. Since the eigenvectors $\mathbf{W}$ exist in the augmented $m \times d$ dimensional space, the spatial \dmd{} modes $\boldsymbol{\Phi}$ are recovered by extracting the first $m$ components (the first block) of the projected eigenvectors:
\begin{equation}
    \boldsymbol{\Phi} = \mathcal{U}_{1:m, :} \mathbf{W}
\end{equation}
Each mode $\phi_i$ in $\boldsymbol{\Phi}$ represents a spatio-temporal pattern, while the corresponding eigenvalue in $\Lambda$ describes the temporal evolution (frequency and growth/decay rate) of that specific pattern.

\section{Experimental setting and detailed results}\label{app:setting}

We fine-tune all the models on a single NVIDIA H100 GPU, and we utilize NVIDIA L40 GPUs for model inference and execution of the input attribution pipelines. 

\subsection{Sentiment analysis}

We use Amazon polarity dataset which has reviews extracted from Amazon and their corresponding labels as \textit{positive} and \textit{negative} \cite{zhang2015character}. All the three LLMs used in this study, were fully fine-tuned on 150,000 train samples in full precision. The test accuracy achieved on 10,000 test samples was around 97\% for all the three models with very minor variations. 

\noindent The instruction prompt used for fine-tuning the LLMs and also used to get the \textit{bias} vector without passing the input sentence for sentiment analysis is as follows.

\begin{figure}[H]
    \centering
    \includegraphics[width=\columnwidth]{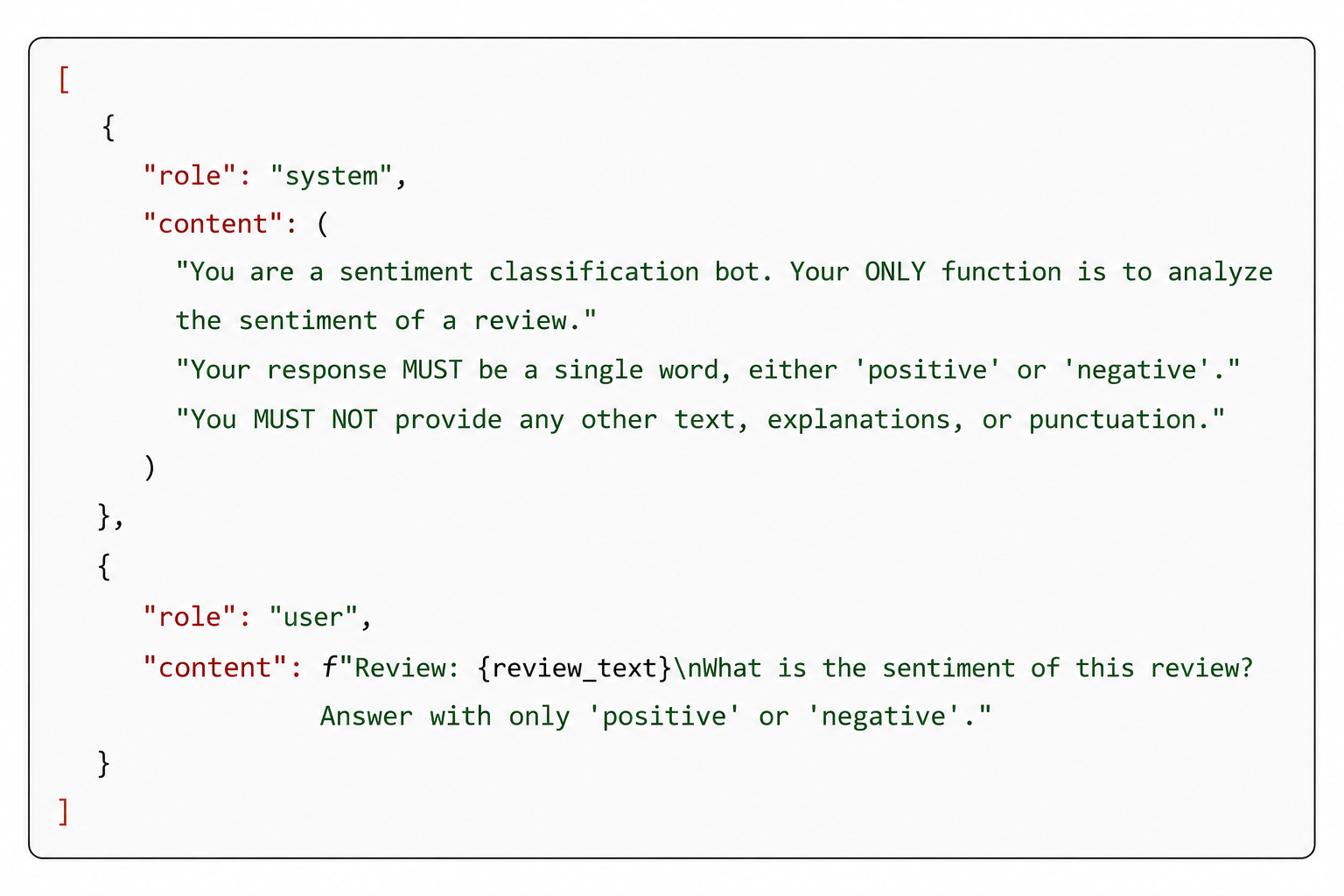}
    \label{fig:instruction_template1}
\end{figure}

\noindent We use the chat template format provided by HuggingFace to define the above chat template, and then HuggingFace handles the tokenization of this template internally. To obtain the vector corresponding to the entire representation of the instruction template, we find the review index \_start and then use the index before that to extract the bias vector from the hidden states. Table \ref{detailed_tables} shows the comparison study of our \dmd{} attribution pipeline of all possible ranking variants against the baselines. From the metrics, it is evident that the modes whose amplitudes persist for longer, capture the sentiment better than those whose initial amplitude has a very high value.  

\subsection{\hx{}}

From \hx{} \cite{mathew2021hatexplain}, we only consider attribution towards hateful sentences, although the model was trained for binary classification with \textit{hateful} and \textit{normal} labels. After fine-tuning on 10,000 samples for 2 epochs, \lama{} achieved an accuracy of 82\%, \qwn{} achieved an accuracy of 83.2\% and \mist{} achieved an accuracy of 84\% on the test set consisting of 2,000 samples. 

\noindent The instruction prompt used for fine-tuning the LLMs is as follows: 
\begin{figure}[H]
    \centering
    \includegraphics[width=\columnwidth]{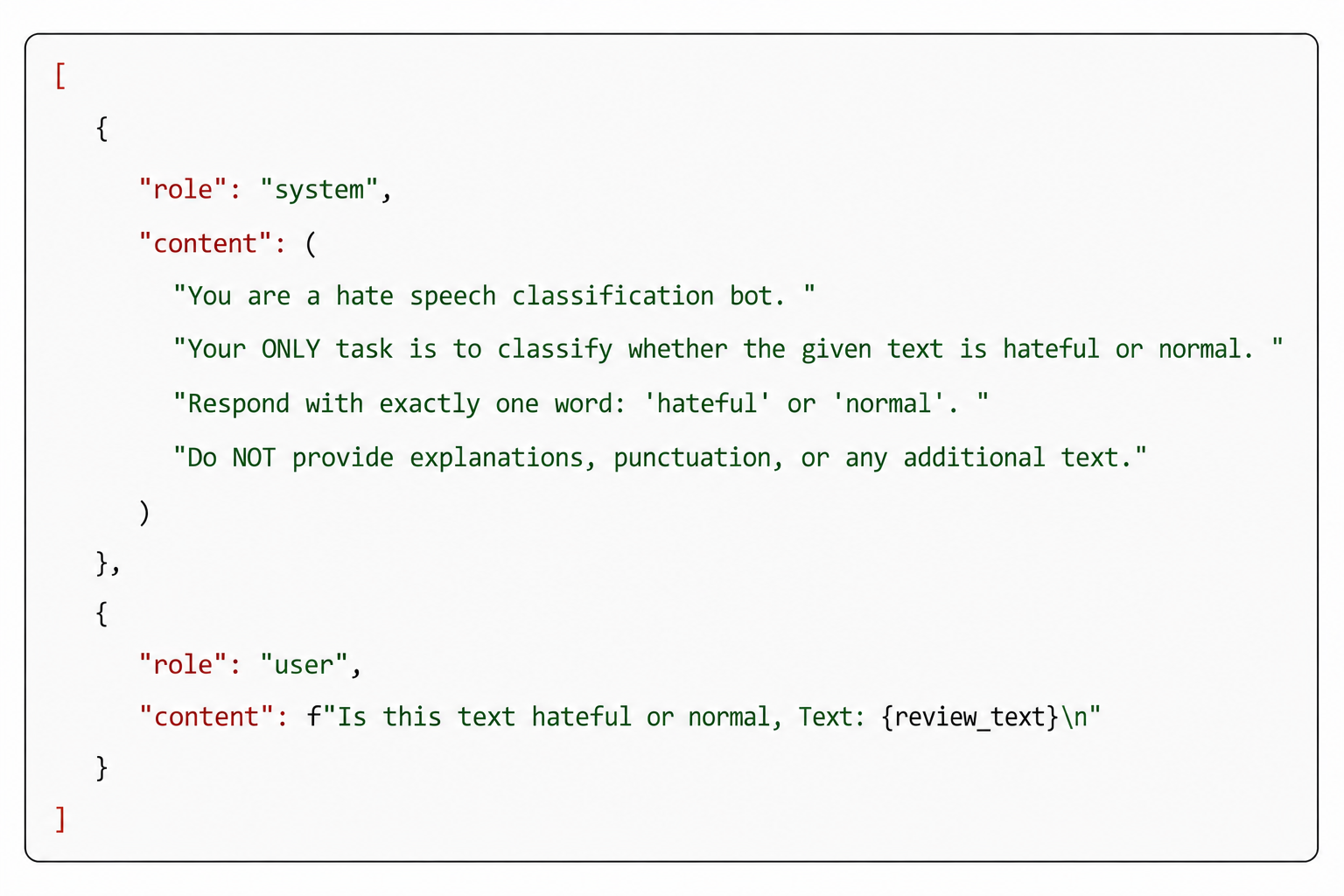}
    \label{fig:instruction_template2}
\end{figure}

\noindent The same prompt was also used for the generation of bias vector from the layer considered in the experiments. Table \ref{detailed_tables} shows full comparison study between different \dmd{} algorithms and ranking techniques. As the dataset contains the hateful triggers as single words or a few collection of words, it can be well approximated by vanilla \dmd{}, and the modes whose amplitude starts high prove to be useful for finding the correct attribution tokens. 

\subsection{\fk{}}

We use \fk{} dataset, which contains posts taken from Reddit and categorized into two labels; \textit{fake} and \textit{not-fake}. These reviews are mostly post titles with short to medium input length sentences. We fine-tune the LLMs on a train set of 15,000 samples for 2 epochs and achieve an accuracy of 84\% for \lama{}, 85.6\% for \qwn{} and 86\% for \mist{} on a test set of 2,000 samples. The prompt template used in fine-tuning and generating the bias vector is as follows. 

\begin{figure}[H]
    \centering
    \includegraphics[width=\columnwidth]{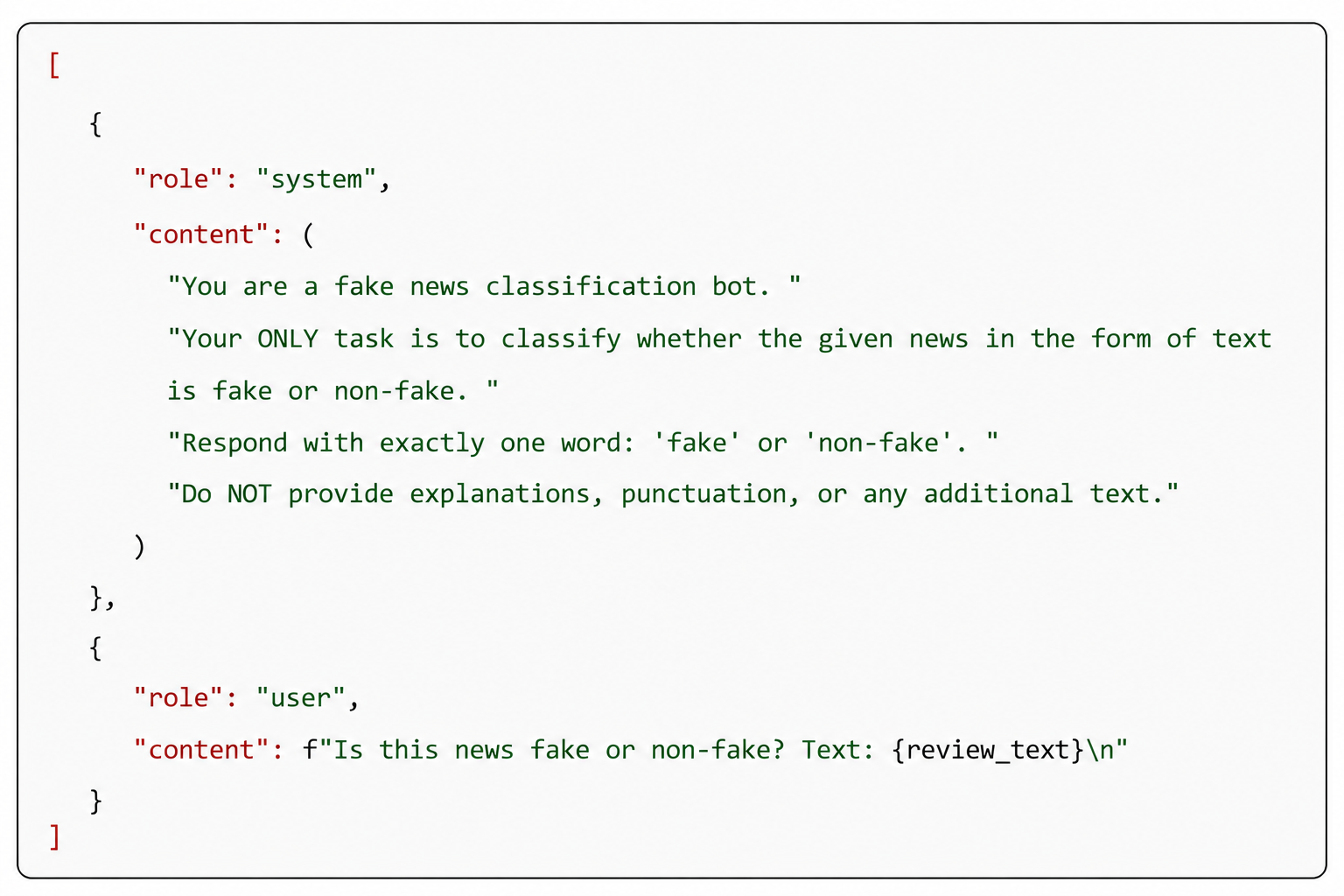}
    \label{fig:instruction_template3}
\end{figure}

\noindent The observations for \fk{} are mostly similar to \hx{}, as \fk{} also contains individual trigger words which can determine the ``fakeness'' of a sentence. Consequently, \dmd{} with amplitude ranking performs the best among all variants and baselines. 

\subsection{\ig{} and \shap{}}

We use Captum library \cite{kokhlikyan2020captum} to implement the pipelines for \ig{} and \shap{}. In both implementations, we calculate the attributions on the embedding layer for each token, which directly measures the impact of the input tokens on the generated output.

\begin{table*}[p]
\centering
\small
\caption{Detailed Attribution results across models and datasets. Best results are noted in \textbf{bold}.}
\label{tab:appendix_full_results}

\begin{minipage}{0.48\textwidth}
\centering
\textbf{\lama{} - \snt{}} \\
\smallskip
\begin{tabularx}{\linewidth}{l|CC|CC}
\toprule
\multirow{2}{*}{\textbf{Method}} & \multicolumn{2}{c|}{\textbf{Negative}} & \multicolumn{2}{c}{\textbf{Positive}} \\
& \textbf{RBO} & \textbf{Rec} & \textbf{RBO} & \textbf{Rec} \\ \midrule
\ig{} & 0.23 & 0.58 & 0.24 & 0.66 \\
\shap{}         & 0.15 & 0.33 & 0.17 & 0.38 \\
\pca{}                  & 0.24 & 0.66 & 0.26 & 0.73 \\ \midrule
\method{}$^{\dmd{}-amp}$ & 0.24 & 0.67 & 0.26 & 0.74 \\
\method{}$^{\dmd{}-avgamp}$ & 0.24 & 0.67 & \textbf{0.27} & 0.75 \\
\method{}$^{\hdmd{}-amp}$ & 0.24 & \textbf{0.68}& \textbf{0.27} & 0.75 \\
\method{}$^{\hdmd{}-avgamp}$  & \textbf{0.25} & \textbf{0.68} & \textbf{0.27} & \textbf{0.76} \\ \bottomrule
\end{tabularx}
\end{minipage}
\hfill
\begin{minipage}{0.48\textwidth}
\centering
\textbf{\qwn{} - \snt{}} \\
\smallskip
\begin{tabularx}{\linewidth}{l|CC|CC}
\toprule
\multirow{2}{*}{\textbf{Method}} & \multicolumn{2}{c|}{\textbf{Negative}} & \multicolumn{2}{c}{\textbf{Positive}} \\
& \textbf{RBO} & \textbf{Rec} & \textbf{RBO} & \textbf{Rec} \\ \midrule
\ig{} & 0.25 & 0.57 & 0.28 & 0.63 \\
\shap{}         & 0.20 & 0.48 & 0.21 & 0.50 \\
\pca{}                  & 0.25 & \textbf{0.69} & 0.28 & 0.74 \\ \midrule
\method{}$^{\dmd{}-amp}$ & 0.24 & 0.68 & 0.27 & 0.74 \\
\method{}$^{\dmd{}-avgamp}$ & 0.24 & 0.68 & 0.28 & \textbf{0.76} \\
\method{}$^{\hdmd{}-amp}$ & 0.24 & 0.68 & 0.27 & 0.75 \\
\method{}$^{\hdmd{}-avgamp}$  & \textbf{0.26} & \textbf{0.69} & \textbf{0.29} & \textbf{0.76} \\ \bottomrule
\end{tabularx}
\end{minipage}

\vspace{2.5em}

\begin{minipage}{0.48\textwidth}
\centering
\textbf{\mist{} - \snt{}} \\
\smallskip
\begin{tabularx}{\linewidth}{l|CC|CC}
\toprule
\multirow{2}{*}{\textbf{Method}} & \multicolumn{2}{c|}{\textbf{Negative}} & \multicolumn{2}{c}{\textbf{Positive}} \\
& \textbf{RBO} & \textbf{Rec} & \textbf{RBO} & \textbf{Rec} \\ \midrule
\ig{} & \textbf{0.27} & 0.67 & 0.26 & 0.61 \\
\shap{}         & 0.20 & 0.48 & 0.22 & 0.51 \\
\pca{}                  & 0.24 & 0.65 & 0.25 & 0.70 \\ \midrule
\method{}$^{\dmd{}-amp}$ & 0.23 & 0.65 & 0.25 & 0.72 \\
\method{}$^{\dmd{}-avgamp}$ & 0.24 & 0.67 & 0.26 & 0.73 \\
\method{}$^{\hdmd{}-amp}$ & 0.24 & 0.66 & 0.26 & 0.74 \\
\method{}$^{\hdmd{}-avgamp}$  & 0.25 & \textbf{0.68} & \textbf{0.27} & \textbf{0.75} \\ \bottomrule
\end{tabularx}
\end{minipage}
\hfill
\begin{minipage}{0.48\textwidth}
\centering
\textbf{\lama{} - \hx{} and \fk{}} \\
\smallskip
\begin{tabularx}{\linewidth}{l|CC|CC}
\toprule
\multirow{2}{*}{\textbf{Method}} & \multicolumn{2}{c|}{\textbf{Hateful}} & \multicolumn{2}{c}{\textbf{Fake}} \\
& \textbf{RBO} & \textbf{Rec} & \textbf{RBO} & \textbf{Rec} \\ \midrule
\ig{} & 0.24 & 0.60 & 0.38 & 0.60 \\
\shap{}         & 0.22 & 0.50 & 0.32 & 0.48 \\
\pca{}                  & 0.23 & 0.60 & \textbf{0.43} & 0.70 \\ \midrule
\method{}$^{\dmd{}-amp}$ & \textbf{0.25} & \textbf{0.64} & 0.42 & \textbf{0.76} \\
\method{}$^{\dmd{}-avgamp}$ & 0.24 & 0.63 & 0.42 & 0.75 \\
\method{}$^{\hdmd{}-amp}$ & \textbf{0.25} & \textbf{0.64} & 0.41 & 0.74 \\
\method{}$^{\hdmd{}-avgamp}$  & 0.24 & 0.63 & 0.41 & 0.74 \\ \bottomrule
\end{tabularx}
\end{minipage}

\vspace{2.5em}

\begin{minipage}{0.48\textwidth}
\centering
\textbf{\qwn{} - \hx{} and \fk{}} \\
\smallskip
\begin{tabularx}{\linewidth}{l|CC|CC}
\toprule
\multirow{2}{*}{\textbf{Method}} & \multicolumn{2}{c|}{\textbf{Hateful}} & \multicolumn{2}{c}{\textbf{Fake}} \\
& \textbf{RBO} & \textbf{Rec} & \textbf{RBO} & \textbf{Rec} \\ \midrule
\ig{} & \textbf{0.26} & 0.60 & 0.38 & 0.60 \\
\shap{}         & 0.22 & 0.50 & 0.34 & 0.52 \\
\pca{}                  & 0.24 & 0.60 & 0.41 & 0.68 \\ \midrule
\method{}$^{\dmd{}-amp}$ & 0.25 & \textbf{0.64} & \textbf{0.42} & \textbf{0.75} \\
\method{}$^{\dmd{}-avgamp}$ & 0.24 & 0.62 & 0.41 & 0.74 \\
\method{}$^{\hdmd{}-amp}$ & 0.25 & 0.63 & 0.41 & 0.74 \\
\method{}$^{\hdmd{}-avgamp}$  & 0.24 & 0.62 & 0.41 & 0.73 \\ \bottomrule
\end{tabularx}
\end{minipage}
\hfill
\begin{minipage}{0.48\textwidth}
\centering
\textbf{\mist{} - \hx{} and \fk{}} \\
\smallskip
\begin{tabularx}{\linewidth}{l|CC|CC}
\toprule
\multirow{2}{*}{\textbf{Method}} & \multicolumn{2}{c|}{\textbf{Hateful}} & \multicolumn{2}{c}{\textbf{Fake}} \\
& \textbf{RBO} & \textbf{Rec} & \textbf{RBO} & \textbf{Rec} \\ \midrule
\ig{} & 0.26 & 0.62 & 0.39 & 0.61 \\
\shap{}         & 0.25 & 0.59 & 0.35 & 0.54 \\
\pca{}                  & \textbf{0.30} & \textbf{0.67} &\textbf{0.44} & 0.72 \\ \midrule
\method{}$^{\dmd{}-amp}$ & 0.26 & 0.65 & 0.41 & \textbf{0.74} \\
\method{}$^{\dmd{}-avgamp}$ & 0.26 & 0.64 & 0.41 & 0.74 \\
\method{}$^{\hdmd{}-amp}$ & 0.25 & 0.63 & 0.40 & 0.73 \\
\method{}$^{\hdmd{}-avgamp}$  & 0.25 & 0.63 & 0.40 & 0.73 \\ \bottomrule
\end{tabularx}
\end{minipage}
\label{detailed_tables}
\end{table*}

\section{Analysis of eigenvalues and the modes}

The temporal evolution of the dynamical system is governed by the discrete-time eigenvalues $\lambda_i$ obtained from the eigendecomposition of the linear operator $\mathbf{A}$. These eigenvalues provide critical information regarding the stability, growth, and oscillatory nature of their corresponding \dmd{} modes $\phi_i$. The position of the eigenvalues relative to the unit circle in the complex plane determines the asymptotic behavior of the modes categorized as follows.

\begin{compactitem}
    \item Steady modes ($|\lambda_i| =1$): Eigenvalues falling exactly on the unit circle correspond to pure oscillations with constant amplitude. These represent the steady-state dynamics of the system.
    \item Stable modes ($|\lambda_i| < 1$): Eigenvalues located within the unit circle represent physically damped or transient dynamics. The closer the eigenvalue is to the origin, the more rapid the decay of the mode as the sequence progresses.
    \item Unstable modes ($|\lambda_i| > 1$): Eigenvalues outside the unit circle indicate exponential growth, often associated with diverging hidden state representations or numerical instabilities in the local flow.
\end{compactitem}

\noindent Based on this theory, we conduct an analysis of the eigenvalues and the dynamics of the hidden states of the representative sentences from the test set of all datasets for each model. We find that for all the sentences, the hidden states produced by the model induce \textbf{decaying} and \textbf{stable} modes. While some modes persist forever without any oscillations (mode 0 in the plots), there exist some modes that decay to zero amplitude before the sentence even finishes, and some modes still oscillate without ever converging to zero amplitude.  

\begin{figure*}[p]
    \centering
    
    \begin{subfigure}[b]{0.48\textwidth}
        \centering
        \includegraphics[width=\textwidth]{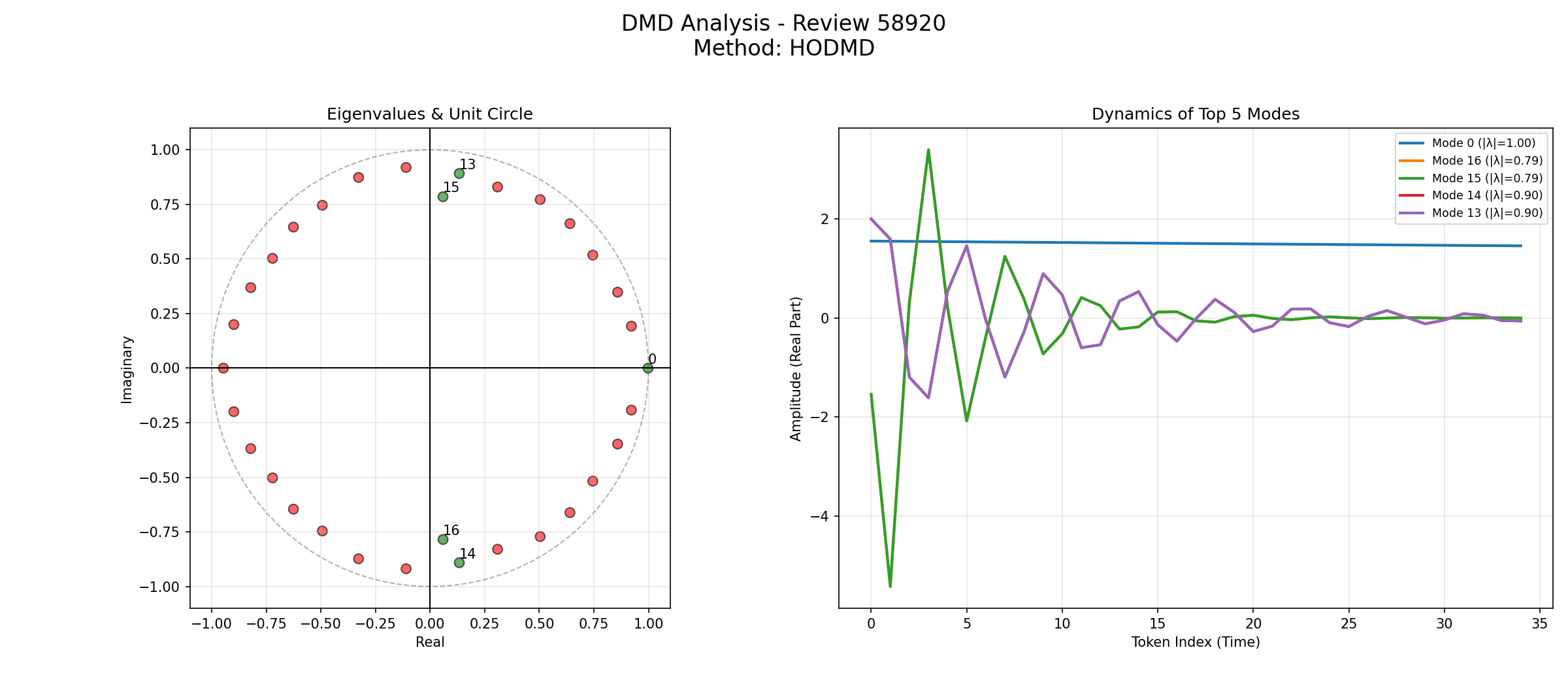}
        \caption{\lama{}: Negative sentiment}
    \end{subfigure}
    \hfill
    \begin{subfigure}[b]{0.48\textwidth}
        \centering
        \includegraphics[width=\textwidth]{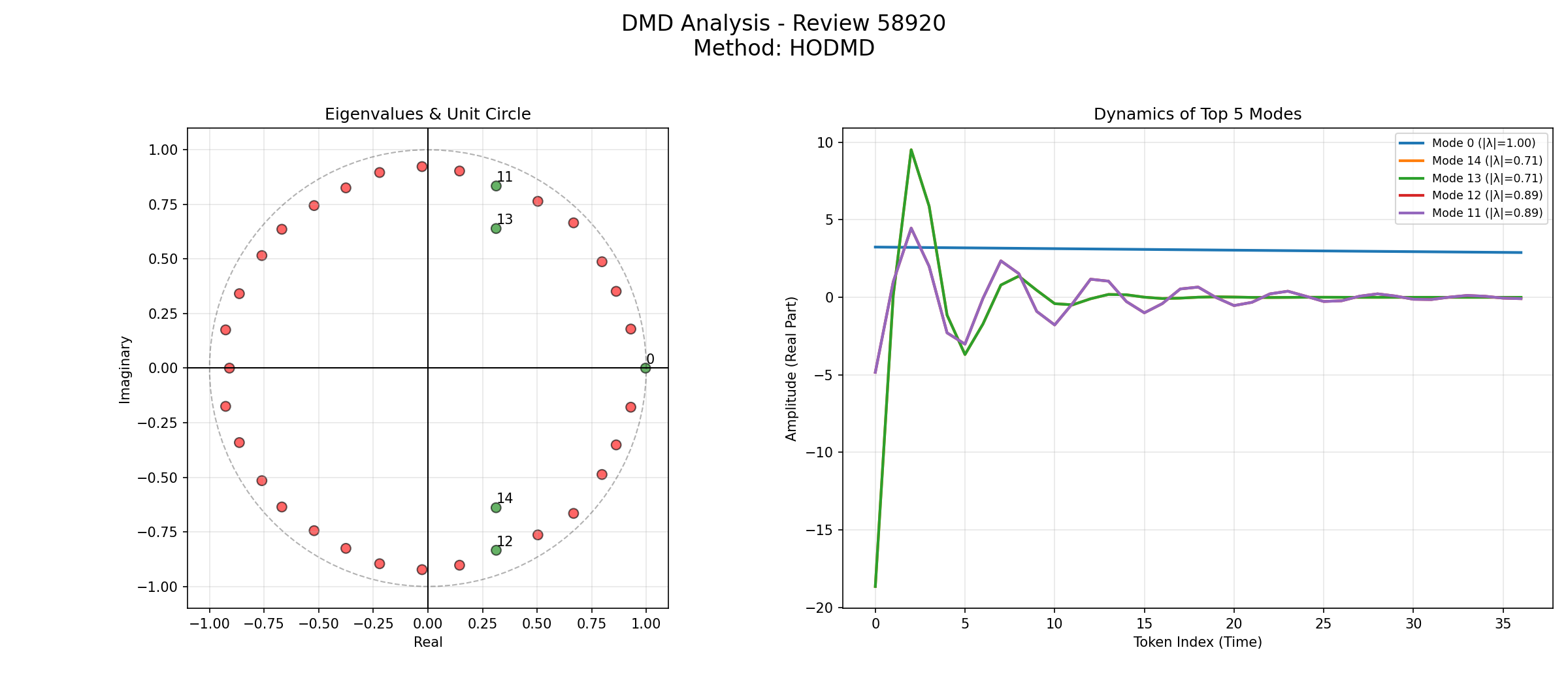}
        \caption{\qwn{}: Negative sentiment}
    \end{subfigure}

    \vspace{0.40em}

    \begin{subfigure}[b]{0.48\textwidth}
        \centering
        \includegraphics[width=\textwidth]{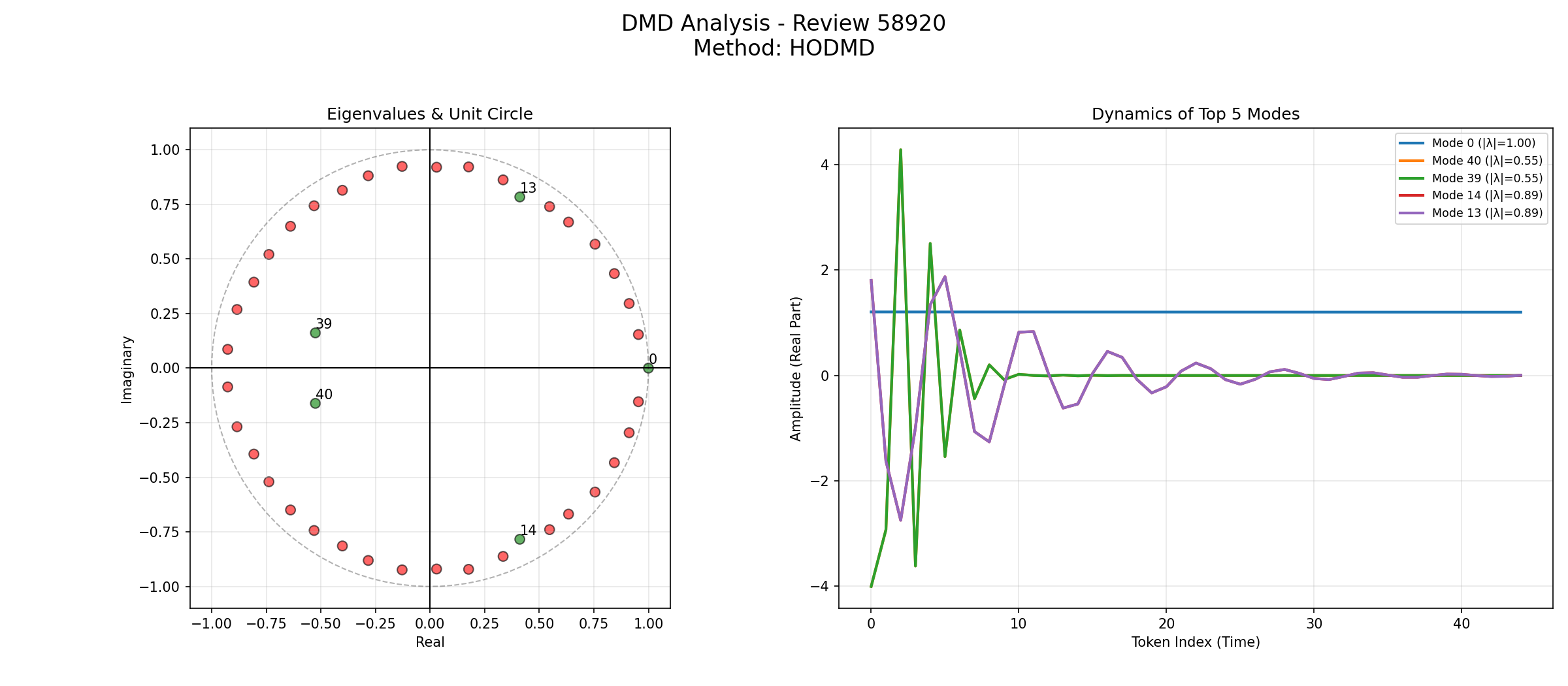}
        \caption{\mist{}: Negative sentiment}
    \end{subfigure}
    \hfill
    \begin{subfigure}[b]{0.48\textwidth}
        \centering
        \includegraphics[width=\textwidth]{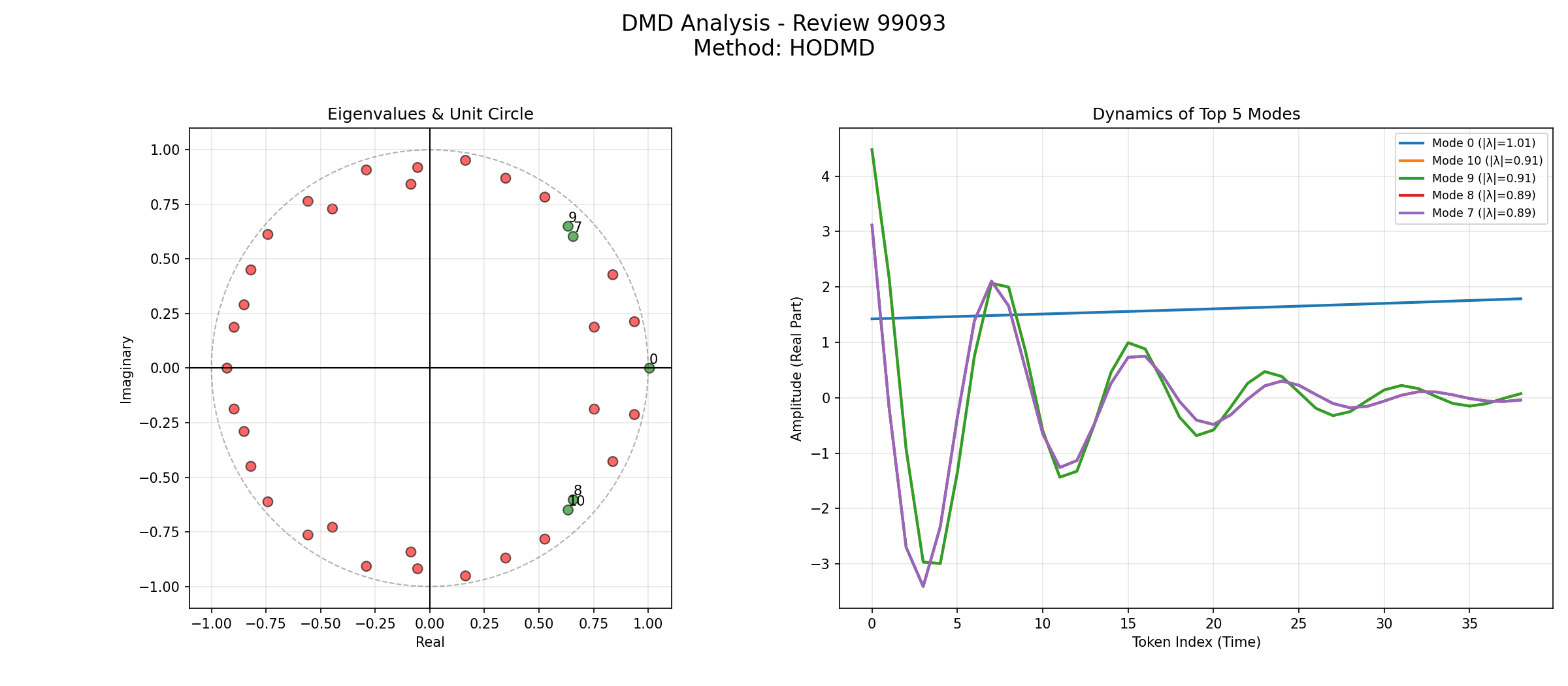}
        \caption{\lama{}: Positive sentiment}
    \end{subfigure}

    \vspace{0.40em}

    \begin{subfigure}[b]{0.48\textwidth}
        \centering
        \includegraphics[width=\textwidth]{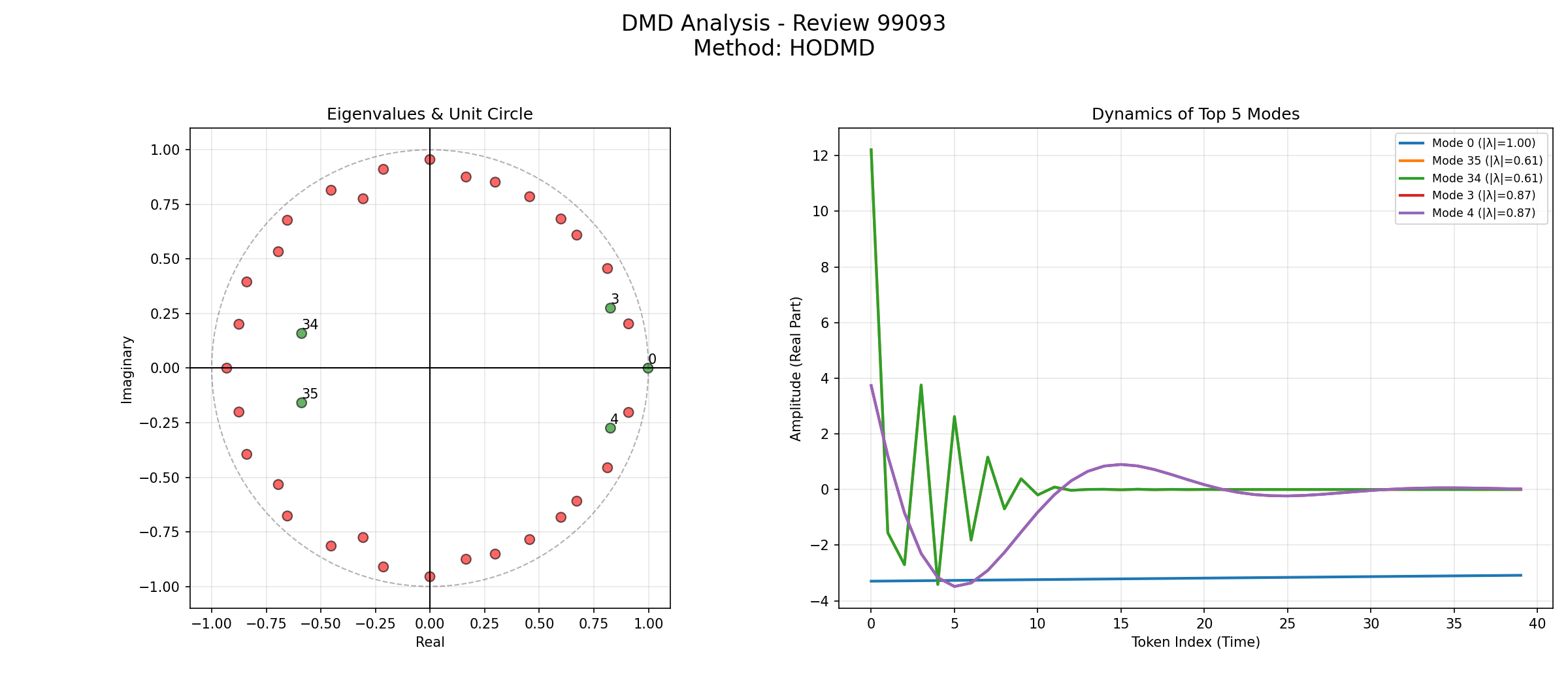}
        \caption{\qwn{}: Positive sentiment}
    \end{subfigure}
    \hfill
    \begin{subfigure}[b]{0.48\textwidth}
        \centering
        \includegraphics[width=\textwidth]{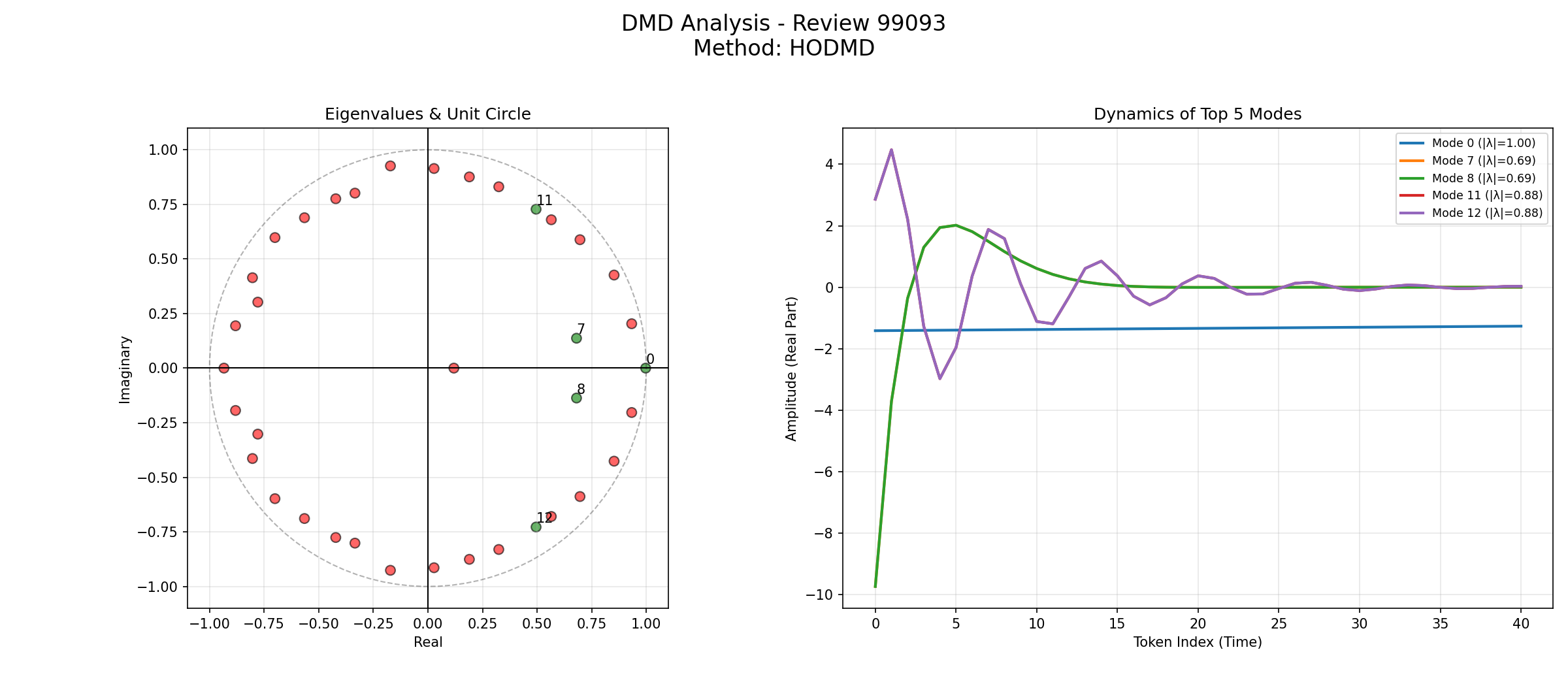}
        \caption{\mist{}: Positive sentiment}
    \end{subfigure}

    \vspace{0.40em}

    \begin{subfigure}[b]{0.48\textwidth}
        \centering
        \includegraphics[width=\textwidth]{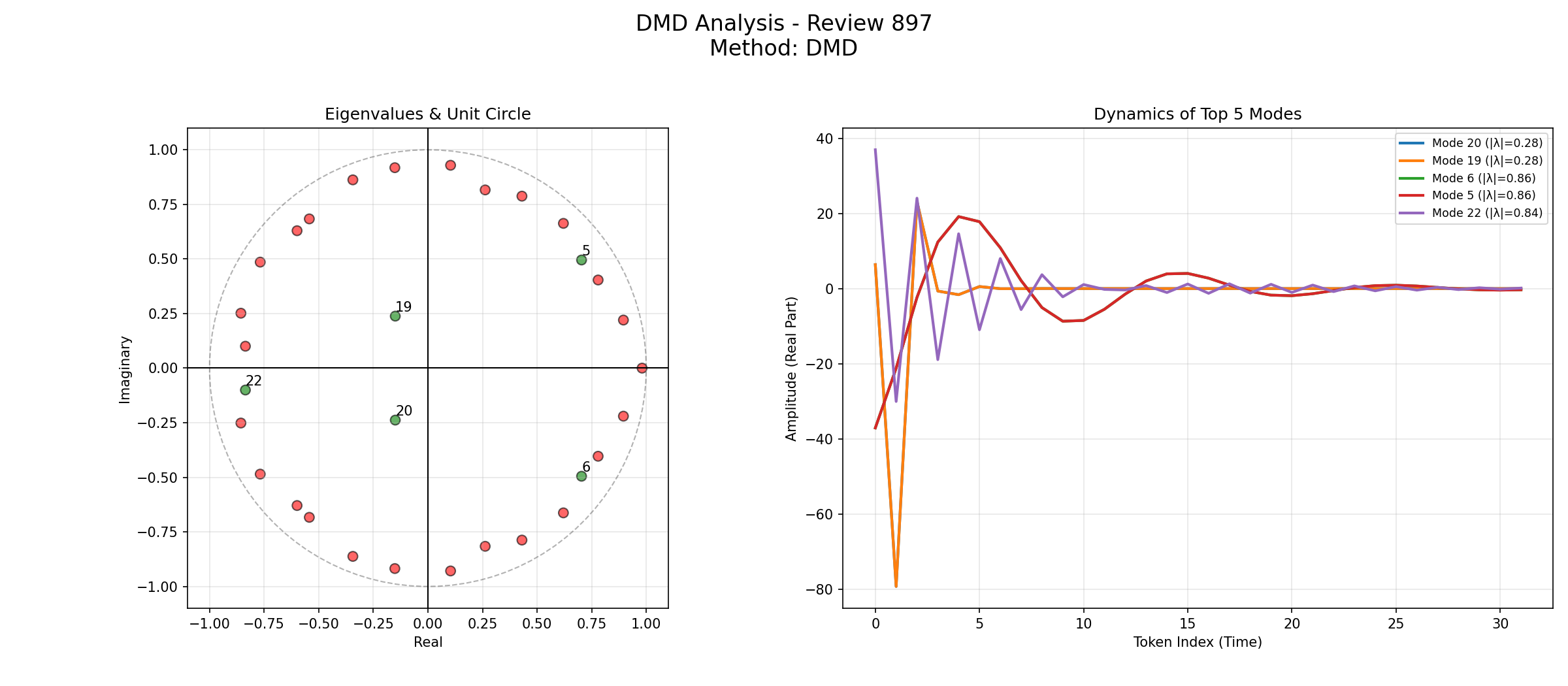}
        \caption{\lama{}: \hx{}}
    \end{subfigure}
    \hfill
    \begin{subfigure}[b]{0.48\textwidth}
        \centering
        \includegraphics[width=\textwidth]{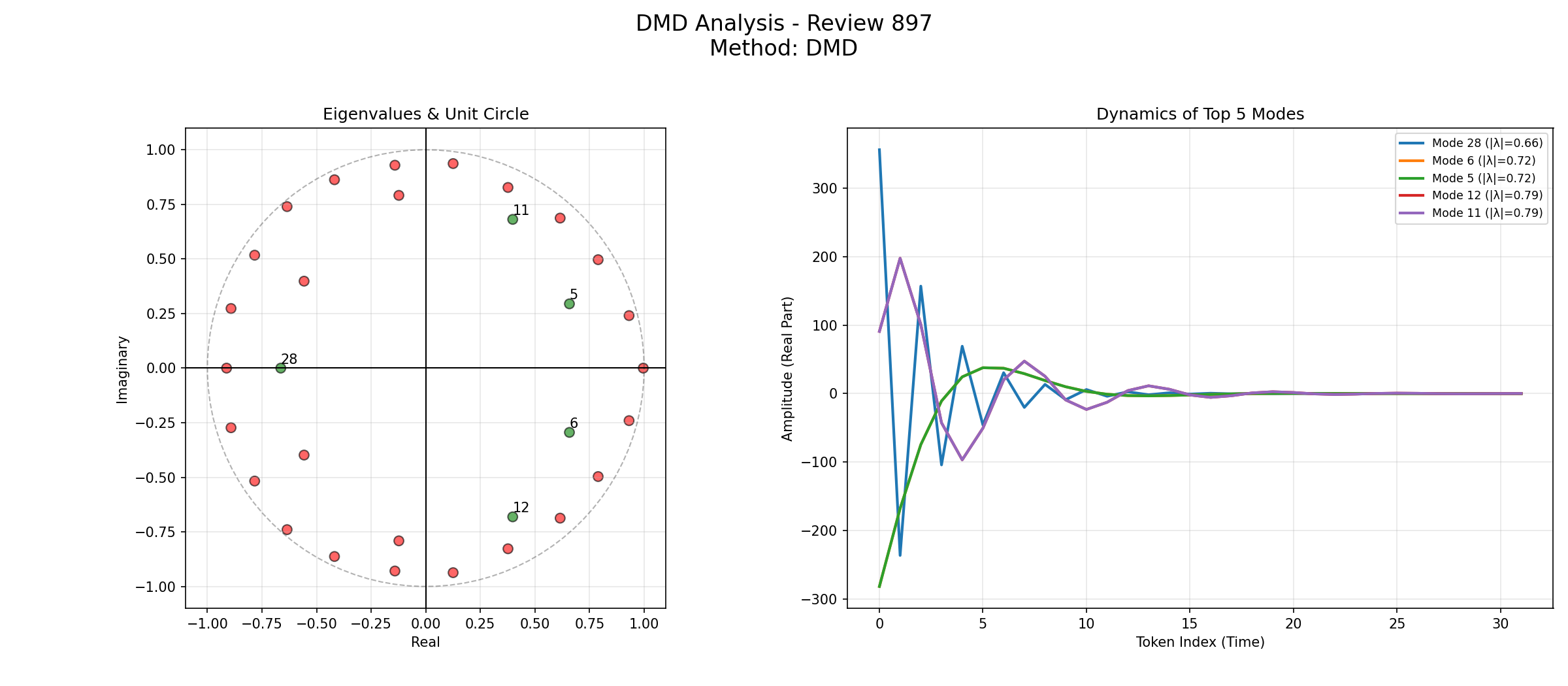}
        \caption{\qwn{}: \hx{}}
    \end{subfigure}

    \vspace{0.40em}

    \begin{subfigure}[b]{0.48\textwidth}
        \centering
        \includegraphics[width=\textwidth]{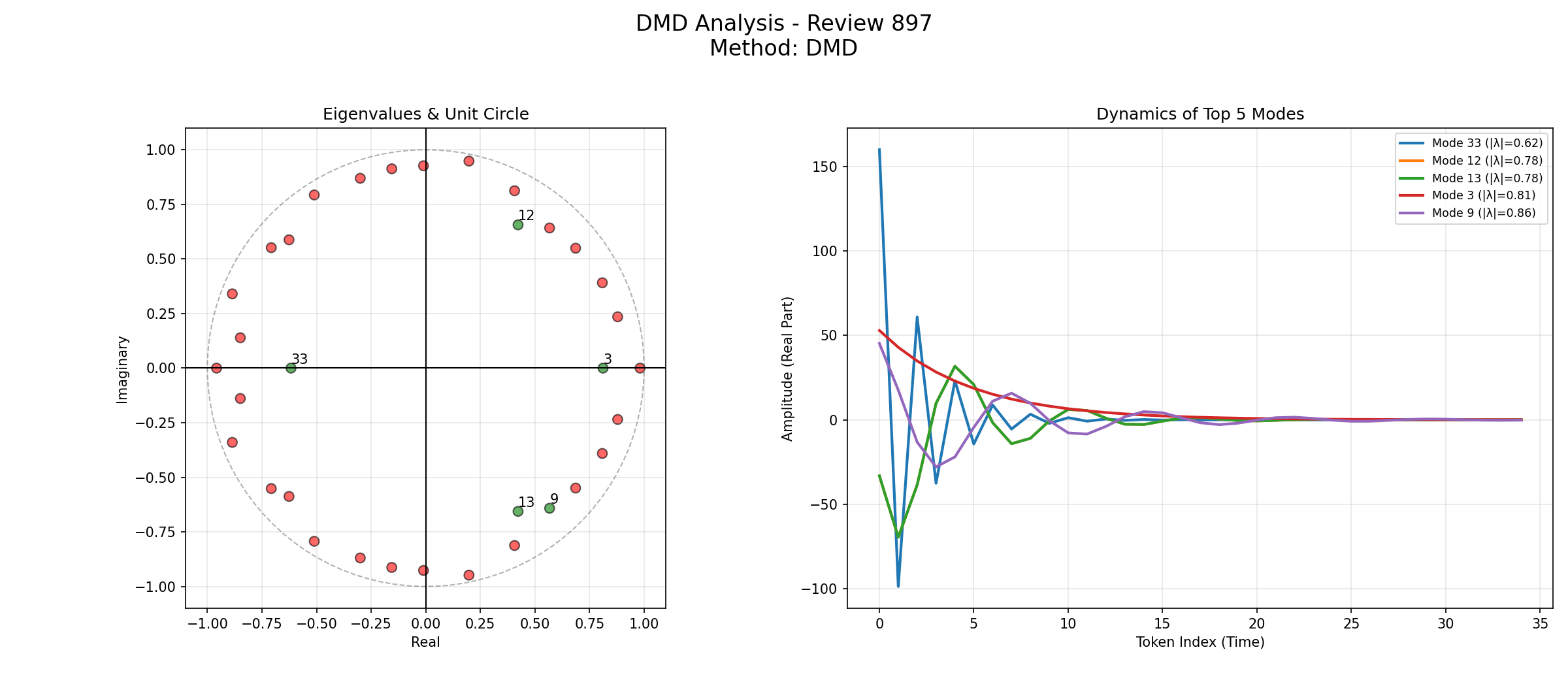}
        \caption{\mist{}: \hx{}}
    \end{subfigure}
    \hfill
    \begin{subfigure}[b]{0.48\textwidth}
        \centering
        \includegraphics[width=\textwidth]{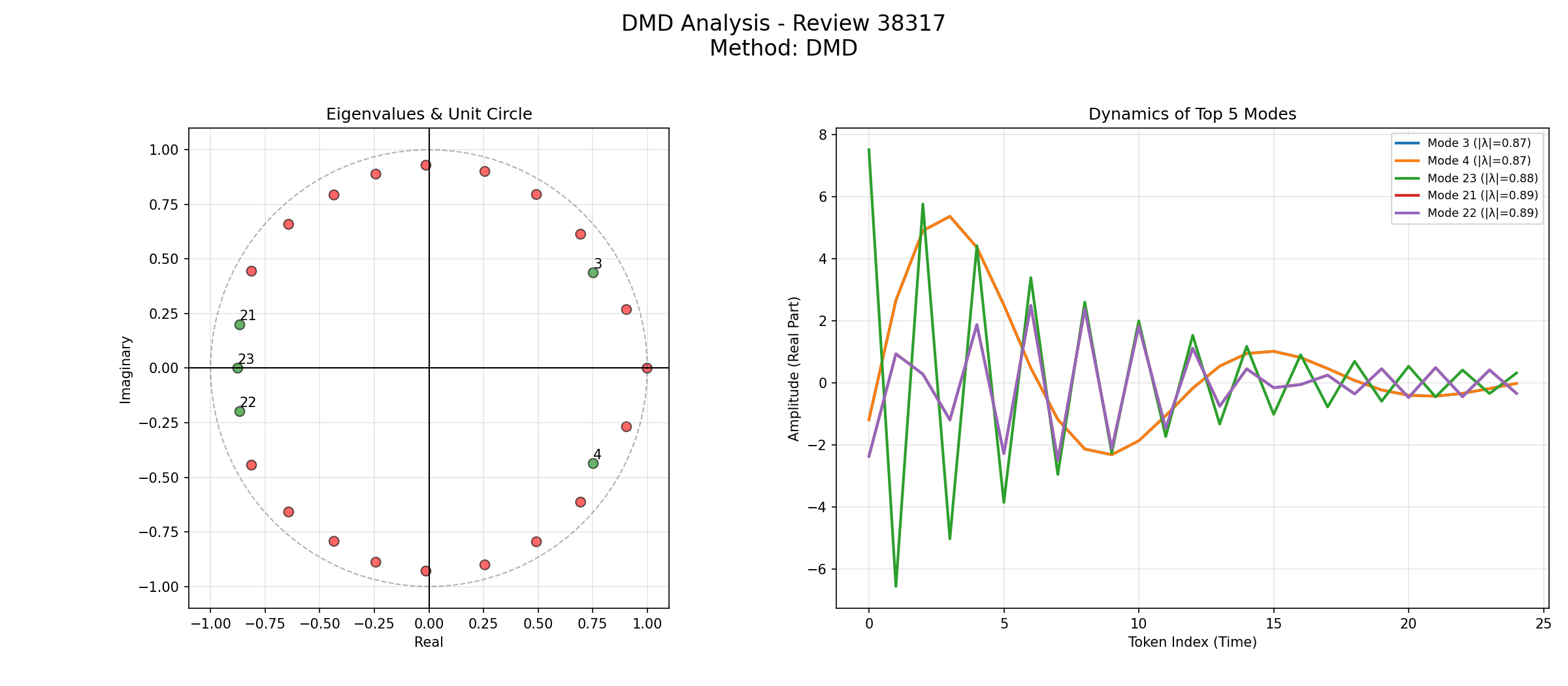}
        \caption{\lama{}: \fk{}}
    \end{subfigure}

    \vspace{0.40em}

    \begin{subfigure}[b]{0.48\textwidth}
        \centering
        \includegraphics[width=\textwidth]{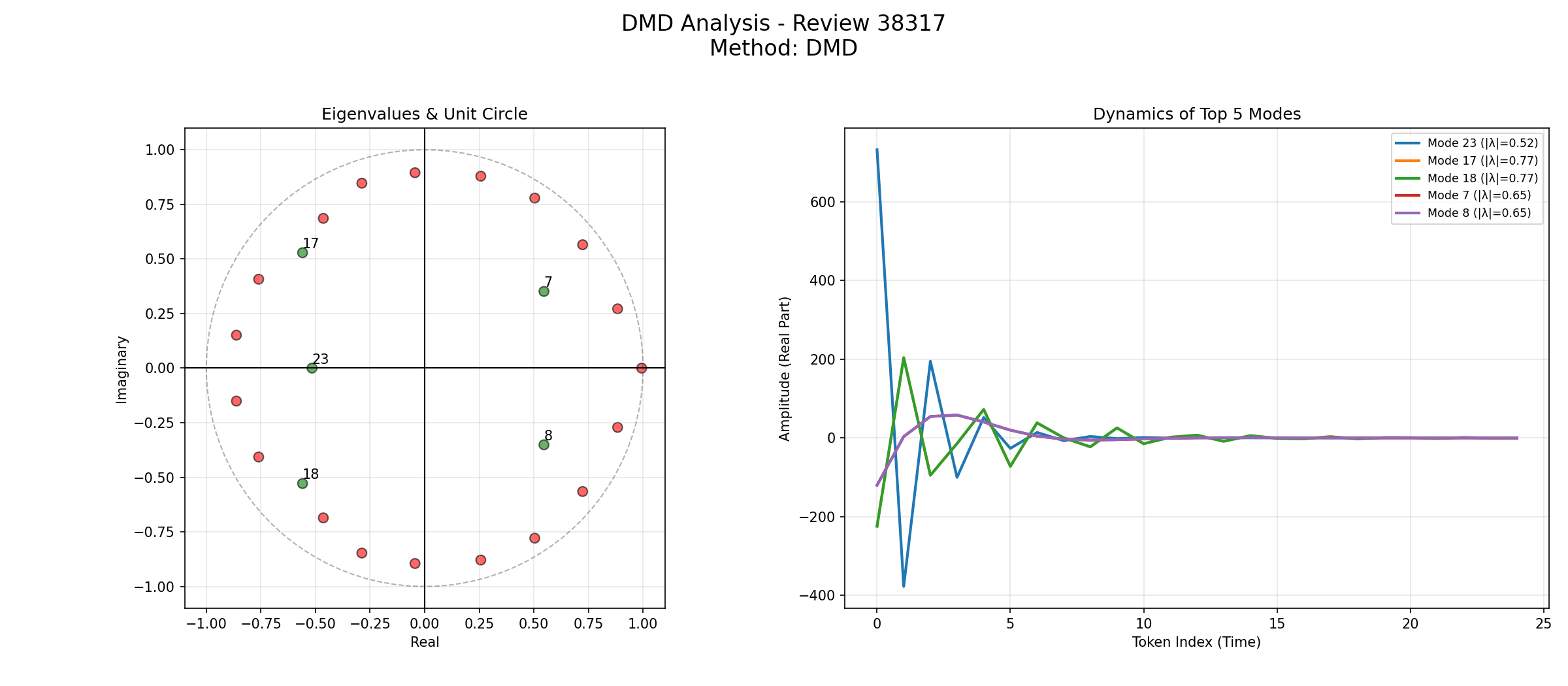}
        \caption{\qwn{}: \fk{}}
    \end{subfigure}
    \hfill
    \begin{subfigure}[b]{0.48\textwidth}
        \centering
        \includegraphics[width=\textwidth]{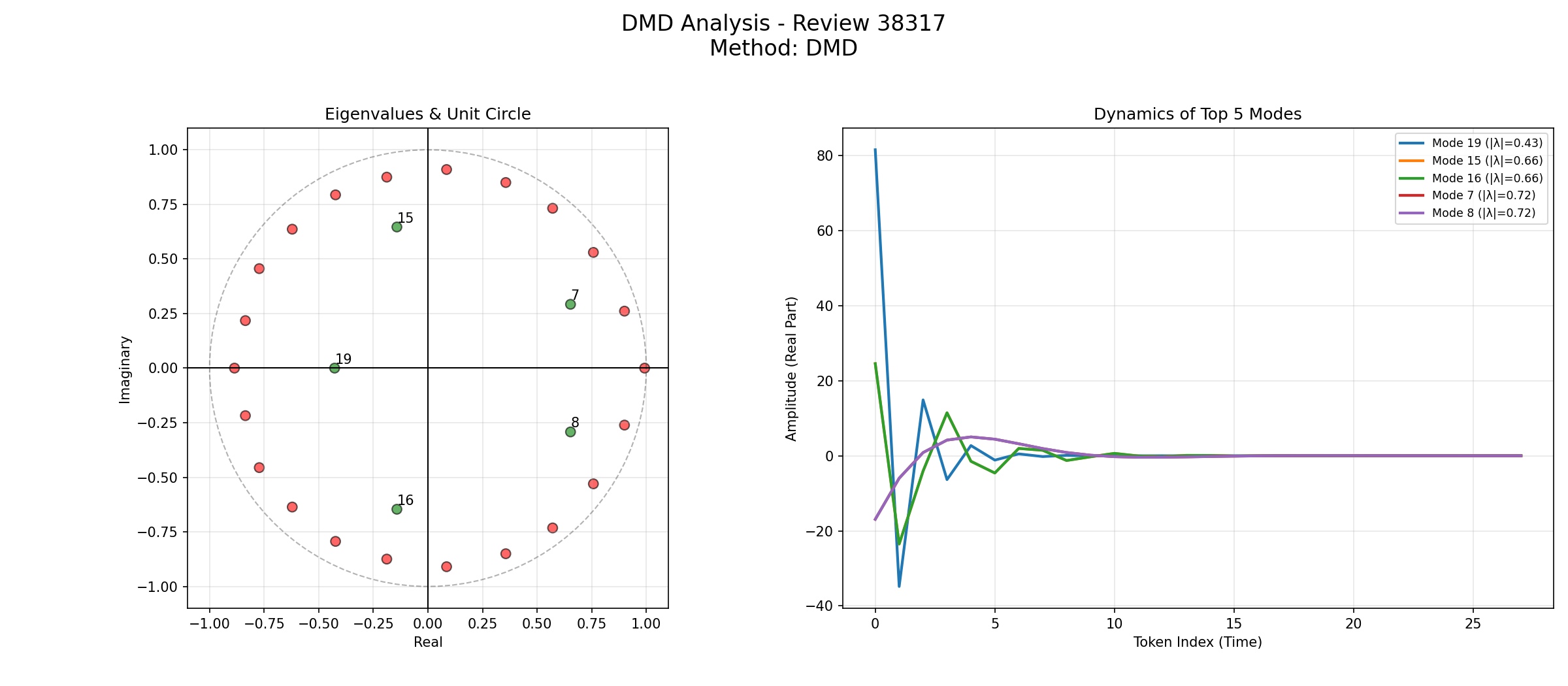}
        \caption{\mist{}: \fk{}}
    \end{subfigure}

    \caption{Eigenvalue distributions and mode dynamics across various model and dataset configurations.}
\label{dmd_analysis_plots}
\end{figure*}

\noindent Figure \ref{dmd_analysis_plots} provides a comprehensive comparison of eigenvalue distributions and mode dynamics across various model and dataset configurations. Each sub-figure consists of two complementary plots: the left plot displays the unit circle (with real and imaginary components on the axes), while the right plot illustrates the corresponding temporal dynamics. In the dynamics plot, the $x$-axis represents the token index (time-step) and the $y$-axis tracks the real part of the mode’s amplitude.
These dynamics plots effectively show how the influence of specific modes fluctuates throughout the processing of a sentence. To ensure a fair comparison, we use the same sample input sentences for every model-dataset pair. The results reveal that even when processing identical inputs, the selected modes and their behaviors vary significantly across models. This divergence is likely due to differences in hidden state dimensions, which force each model to encode and process information in its own distinct way. This observation is further supported by the results in Table \ref{detailed_tables} of this Appendix. Although \pca{} performance fluctuates significantly between different models even on the same dataset, our optimized \dmd{} configuration remains remarkably stable. This consistency suggests that \dmd{} is less sensitive to architectural variations, providing more reliable attribution metrics across diverse model scales and hidden state dimensions.

\section{Error analysis}

We also conduct an error analysis of how good is the approximation $\mathbf{A}$ found by \dmd{} by reconstructing the hidden states for sample sentences and then measuring the Frobenius norm between the original hidden states and the reconstructed states.
The reconstruction of the snapshot matrix $\hat{\bm{X}}_\text{orig}$ is achieved by utilizing the extracted \dmd{} modes $\boldsymbol{\Phi}$, the diagonal matrix of eigenvalues $\bm{\Lambda}$, and the vector of initial amplitudes $\boldsymbol{\alpha}$. The approximated state at time $t$ is given by:

\begin{equation}
    \hat{h}_t = \sum_{i=1}^{r} \phi_i \lambda_i^{t-1} \alpha_i = \boldsymbol{\Phi} \bm{\Lambda}^{t-1} \boldsymbol{\alpha}
\end{equation}

\noindent For the entire sequence of $N$ snapshots, the reconstructed matrix $\hat{\bm{X}}_\text{recons}$ is represented as:

\begin{equation}
    \hat{\bm{X}}_\text{recons} = \boldsymbol{\Phi} \text{diag}(\boldsymbol{\alpha}) \begin{bmatrix} 
    1 & \lambda_1 & \dots & \lambda_1^{N-1} \\
    1 & \lambda_2 & \dots & \lambda_2^{N-1} \\
    \vdots & \vdots & \ddots & \vdots \\
    1 & \lambda_r & \dots & \lambda_r^{N-1} 
    \end{bmatrix}
\end{equation}

\noindent To evaluate the fidelity of this approximation, we calculate the relative reconstruction error using the Frobenius norm $||\cdot||_F$. The error $\epsilon$ is defined as the ratio of the norm of the residual to the norm of the original snapshot matrix:

\begin{equation}
    \epsilon = \frac{||\hat{\bm{X}}_\text{orig} - \hat{\bm{X}}_\text{recons}||_F}{||\hat{\bm{X}}_\text{orig}||_F}
\label{error_term}
\end{equation}

Table \ref{reconstruction_err} represents the error values for the models and the datasets. We find that the reconstruction error rate is very low in the order of $10^{-2}$, suggesting that the linear approximation of the LLM using \dmd{} is quite accurate. 

\begin{table*}[t]
\centering
\small
\renewcommand{\arraystretch}{1.3}

\begin{tabularx}{\textwidth}{
    >{\raggedright\arraybackslash}X
    >{\centering\arraybackslash}X
    >{\centering\arraybackslash}X
    >{\centering\arraybackslash}X
}
\toprule
\multirow{2}{*}{\textbf{Label}} &
\multicolumn{3}{c}{\textbf{Reconstruction error ($\epsilon$)}} \\
\cmidrule(lr){2-4}
& \lama{} & \qwn{} & \mist{} \\
\midrule
Negative Sentiment & 0.070 & 0.074 & 0.059 \\
Positive Sentiment & 0.069 & 0.074 & 0.060 \\
Hateful Label      & 0.096 & 0.073 & 0.059 \\
Fake Label         & 0.132 & 0.099 & 0.121 \\
\bottomrule
\end{tabularx}

\caption{Reconstruction error $\epsilon$.}
\label{reconstruction_err}
\end{table*}

\begin{table*}[t]
\centering
\scriptsize
\setlength{\tabcolsep}{3pt}

\begin{tabularx}{\textwidth}{l | l | *{2}{C} | *{2}{C} | *{2}{C}}
\toprule
\multirow{2}{*}{\textbf{Model}} & \multirow{2}{*}{\textbf{Method}} & \multicolumn{2}{c|}{\textbf{Sentiment Analysis}} & \multicolumn{2}{c|}{\textbf{HateXplain}} & \multicolumn{2}{c}{\textbf{Fakeddit}} \\ 
\cmidrule(lr){3-4} \cmidrule(lr){5-6} \cmidrule(lr){7-8}
& & \textbf{GPU (GB) $\downarrow$} & \textbf{Time (s) $\downarrow$} & \textbf{GPU (GB) $\downarrow$} & \textbf{Time (s) $\downarrow$} & \textbf{GPU (GB) $\downarrow$} & \textbf{Time (s) $\downarrow$} \\ 
\midrule

\multirow{4}{*}{\textbf{\lama{}}} 
& IG & 15.23 $\pm$ 1.20 & 9.07 / 8.24 & 14.70 $\pm$ 0.50 & 6.17 & 14.18 $\pm$ 0.12 & 5.86 \\
& SHAP & 38.80 $\pm$ 3.00 & 2.62 / 2.30 & 35.30 & 2.21 & 27.22 $\pm$ 4.20 & 1.66 \\
& PCA & \textbf{12.90} & 1.10 / 1.14 & 12.99 & 0.87 & \textbf{12.96} & 0.73 \\
& \method{} (Ours) & 13.01 & \textbf{1.03 / 1.10} & \textbf{12.90} & \textbf{0.33} & 12.97 & \textbf{0.36} \\ 
\midrule

\multirow{4}{*}{\textbf{\qwn{}}} 
& IG & 18.20 $\pm$ 0.90 & 10.98 / 10.02 & 18.15 $\pm$ 0.30 & 5.96 & 17.78 & 5.74 \\
& SHAP & 45.30 $\pm$ 4.20 & 3.37 / 3.56 & 39.56 $\pm$ 3.20 & 2.34 & 40.15 $\pm$ 2.1 & 1.97 \\
& PCA & \textbf{16.18} & 1.61 / 1.65 & \textbf{16.07} & 1.00 & \textbf{16.22} & 0.79 \\
& \method{} (Ours) & 16.31 $\pm$ 0.10 & \textbf{0.87 / 0.96} & 16.13 & \textbf{0.40} & \textbf{16.22} & \textbf{0.37} \\ 
\midrule

\multirow{4}{*}{\textbf{\mist{}}} 
& IG & 28.26 $\pm$ 1.30 & 16.30 / 16.82 & 30.09 $\pm$ 0.20 & 11.40 & 29.86 $\pm$ 0.20 & 11.30 \\
& SHAP & -- & -- & -- & -- & -- & -- \\
& PCA & \textbf{28.26} & 1.73 / 2.21 & \textbf{28.22} & 1.04 & \textbf{28.22} & 0.94 \\
& \method{} (Ours) & 28.28 & \textbf{1.03 / 1.65} & \textbf{28.22} & \textbf{0.51} & \textbf{28.22} & \textbf{0.47} \\ 
\bottomrule
\end{tabularx}

\caption{Detailed efficiency evaluation comparing peak GPU memory usage (GB) and per-sample attribution computation time (seconds) across model families and benchmark datasets. For \snt{} dataset, execution times are reported for negative/positive context passes.}
\label{tab:full_computational_efficiency}
\end{table*}

\section{Computational efficiency and resource utilization}
To assess the computational efficiency of our proposed framework, we benchmark its average execution time per input and peak GPU memory footprint against established baselines. All timing and memory profiling experiments were conducted on NVIDIA L40 GPUs using single-precision floating-point format (\texttt{float32}) for model parameters and internal vector representations. As detailed in Table \ref{tab:full_computational_efficiency}, our approach achieves superior efficiency, requiring the lowest processing time and minimal GPU memory overhead. This performance gain primarily stems from our architectural design: while standard gradient- and perturbation-based techniques like \ig{} and \shap{} require dozens to hundreds of forward and backward passes per input sequence, our method extracts attributions non-intrusively via a single forward pass. Consequently, this yields a substantial reduction in inference latency and resource consumption.

\section{Ablation studies}

\subsection{Results without denoising}
To assess the importance of removing the noise inflicted by the empty chat prompt template, we studied the metrics without the removal. We find that the metrics drop significantly for sentiment analysis, there is a slight drop for \hx{} but the drop in metrics is not so significant for \fk{}. Detailed results for the three models and the datasets for \method{} without the noise removal are presented in the Tables \ref{tab:sentiment_results_with_noise} and \ref{tab:hatexplain_fakeddit_our_method_with_noise}. This reinforces that the noise removal step not only removes the noise, but also ensures that the hidden state snapshots contain the information which the specific token adds to the residual stream.

\begin{table*}[t]
\centering
\scriptsize
\setlength{\tabcolsep}{3pt}
\begin{tabularx}{\textwidth}{l | *{3}{C} || *{3}{C}}
\toprule
\multirow{2}{*}{\textbf{Model}} & \multicolumn{3}{c||}{\textbf{Negative Sentiment}} & \multicolumn{3}{c}{\textbf{Positive Sentiment}} \\ 
\cmidrule(lr){2-4} \cmidrule(lr){5-7}
& \textbf{MC@20 $\uparrow$} & \textbf{RBO@20 $\uparrow$} & \textbf{Recall@20 $\uparrow$} & \textbf{MC@20 $\uparrow$} & \textbf{RBO@20 $\uparrow$} & \textbf{Recall@20 $\uparrow$} \\ 
\midrule
\lama{} & 5.19 & 0.23 & 0.66 & 5.92 & 0.26 & 0.73 \\
\qwn{}  & 5.28 & 0.23 & 0.67 & 5.99 & 0.26 & 0.73 \\
\mist{} & 5.13 & 0.23 & 0.64 & 5.81 & 0.25 & 0.71 \\
\bottomrule
\end{tabularx}
\caption{Evaluation results of our proposed method (\method{}$^{\hdmd{}-avgamp}$) on \snt{} dataset across Negative and Positive sentiment classes for three model families without denoising the hidden states.}
\label{tab:sentiment_results_with_noise}
\end{table*}

\begin{table*}[t]
\centering
\scriptsize
\setlength{\tabcolsep}{3pt}
\begin{tabularx}{\textwidth}{l | *{3}{C} || *{3}{C}}
\toprule
\multirow{2}{*}{\textbf{Model}} & \multicolumn{3}{c||}{\textbf{Hateful Reviews (\hx{})}} & \multicolumn{3}{c}{\textbf{Fake Reviews (\fk{})}} \\ 
\cmidrule(lr){2-4} \cmidrule(lr){5-7}
& \textbf{MC@10 $\uparrow$} & \textbf{RBO@10 $\uparrow$} & \textbf{Recall@10 $\uparrow$} & \textbf{MC@10 $\uparrow$} & \textbf{RBO@10 $\uparrow$} & \textbf{Recall@10 $\uparrow$} \\ 
\midrule
\lama{} & 2.76 & 0.24 & 0.63 & 5.45 & 0.42 & 0.75 \\
\qwn{}  & 2.80 & 0.24 & 0.63 & 5.37 & 0.41 & 0.75 \\
\mist{} & 2.80 & 0.26 & 0.64 & 5.27 & 0.41 & 0.74 \\
\bottomrule
\end{tabularx}
\caption{Evaluation results of our proposed method (\method{}$^{\dmd{}-amp}$) on the \hx{} and \fk{} datasets across three model families without denoising the hidden states.}
\label{tab:hatexplain_fakeddit_our_method_with_noise}
\end{table*}

\subsection{Threshold $\tau=0.5$}
Raising the similarity threshold to $0.5$ restricts layer selection strictly to representations that exhibit a higher alignment with the model's instruction-following dynamics. Intuitively, enforcing a stricter cutoff filters out lower-level contextual signals, which is expected to yield a corresponding decline in attribution performance metrics. Our empirical observations largely align with this expectation, albeit with domain-specific nuances. For the \fk{} dataset, we observe a minor reduction across evaluation metrics. On \hx{}, performance remains largely stable, accompanied by a subtle increase in RBO scores. In contrast, for the \snt{} dataset across both positive and negative sentiment classes, raising the threshold induces a pronounced drop in metric values for \qwn{} and \mist{}, whereas \lama{} maintains relatively resilient performance. Complete results across all configurations are reported in Tables \ref{tab:sentiment_05} and \ref{tab:fakeddit_hatexplain_05}.

\begin{table*}[t]
\centering
\scriptsize
\setlength{\tabcolsep}{3pt}
\begin{tabularx}{\textwidth}{l | *{3}{C} || *{3}{C}}
\toprule
\multirow{4}{*}{\textbf{Method}} & \multicolumn{6}{c}{\textbf{Top20}} \\
\cmidrule(lr){2-7}
& \multicolumn{3}{c||}{\textbf{Negative}} & \multicolumn{3}{c}{\textbf{Positive}} \\
\cmidrule(lr){2-4} \cmidrule(lr){5-7}
& \textbf{Matched count} & \textbf{RBO} & \textbf{Recall@k} & \textbf{Matched count} & \textbf{RBO} & \textbf{Recall@k} \\
\midrule

\multicolumn{7}{c}{\textbf{\lama{}}} \\
\midrule
\method{}$^{\hdmd{}-avgamp}$ & 5.29 & 0.24 & 0.67 & 6.02 & 0.26 & 0.74 \\
PCA                       & 5.31 & 0.25 & 0.67 & 5.91 & 0.26 & 0.72 \\
\midrule

\multicolumn{7}{c}{\textbf{\qwn{}}} \\
\midrule
\method{}$^{\hdmd{}-avgamp}$ & 5.18 & 0.23 & 0.65 & 5.83 & 0.25 & 0.72 \\
PCA                       & 5.09 & 0.24 & 0.64 & 5.60 & 0.23 & 0.69 \\
\midrule

\multicolumn{7}{c}{\textbf{\mist{}}} \\
\midrule
\method{}$^{\hdmd{}-avgamp}$ & 5.16 & 0.23 & 0.64 & 5.78 & 0.25 & 0.71 \\
PCA                       & 5.00 & 0.23 & 0.62 & 5.46 & 0.23 & 0.68 \\
\bottomrule
\end{tabularx}
\caption{Comparative evaluation results for Sentiment Analysis across negative and positive classes with threshold $\tau=0.5$.}
\label{tab:sentiment_05}
\end{table*}

\subsection{Varying the $k$ value in top-$k$ tokens for calculation of metrics}
We check the consistency of the results when we select top-$10$ ranked tokens with the GT for the \snt{} dataset, and top-$20$ ranked tokens with GT for \fk{} and \hx{} datasets. We find that the results remain consistent in the \snt{} dataset, while in that of \fk{} and \hx{} the metrics start to converge for top-$20$ ranked tokens as there are very few input sentences from these datasets which have GT tokens more than 10. Detailed results are presented in Tables \ref{senti_top10} and \ref{hx_fk_top20}.

\begin{table*}[t]
\centering
\scriptsize
\begin{tabularx}{\textwidth}{l|*{3}{C|}|*{3}{C|}}
\toprule
\multirow{2}{*}{\textbf{Method}} & \multicolumn{3}{c|}{\textbf{Negative sentiment}} & \multicolumn{3}{c}{\textbf{Positive sentiment}} \\ 
\cmidrule(lr){2-4} \cmidrule(lr){5-7}
& \textbf{MC@10 $\uparrow$} & \textbf{RBO@10 $\uparrow$} & \textbf{Recall@10 $\uparrow$} & \textbf{MC@10 $\uparrow$} & \textbf{RBO@10 $\uparrow$} & \textbf{Recall@10 $\uparrow$} \\ 
\midrule
\multicolumn{7}{c}{\lama{}} \\ 
\midrule
\ig{} & 2.93 & 0.26 & 0.39 & 3.09 & 0.28 & 0.45 \\
\shap{} & 2.52 & 0.21 & 0.33 & 2.70 & 0.23 & 0.38 \\
\pca{} & 3.19 & 0.26 & 0.42 & 3.51 & 0.28 & 0.45 \\
\method{}$^{\hdmd{}-avgamp}$ & 3.28 & 0.26 & 0.43 & 3.73 & 0.29 & 0.48 \\ 
\midrule

\multicolumn{7}{c}{\qwn{}} \\ 
\midrule
\ig{} & 3.17 & 0.29 & 0.41 & 3.55 & 0.31 & 0.45 \\
\shap{} & 2.73 & 0.22 & 0.35 & 3.00 & 0.24 & 0.37 \\
\pca{} & 3.35 & 0.27 & 0.44 & 3.70 & 0.30 & 0.48 \\
\method{}$^{\hdmd{}-avgamp}$ & 3.43 & 0.28 & 0.45 & 3.80 & 0.31 & 0.49 \\
\midrule

\multicolumn{7}{c}{\mist{}} \\ 
\midrule
\ig{} & 3.41 & 0.31 & 0.43 & 3.45 & 0.29 & 0.43 \\
\shap{} & 2.77 & 0.23 & 0.35 & 3.01 & 0.24 & 0.37 \\
\pca{} & 3.13 & 0.25 & 0.40 & 3.26 & 0.26 & 0.41 \\
\method{}$^{\hdmd{}-avgamp}$ & 3.40 & 0.27 & 0.44 & 3.65 & 0.29 & 0.47 \\
\bottomrule
\end{tabularx}
\caption{Results for top-$10$ ranked tokens in case of \snt{} dataset.}
\label{senti_top10}
\end{table*}

\begin{table*}[t]
\centering
\scriptsize
\begin{tabularx}{\textwidth}{l|*{3}{C|}|*{3}{C|}}
\toprule
\multirow{2}{*}{\textbf{Method}} & \multicolumn{3}{c|}{\textbf{Hateful reviews (\hx{})}} & \multicolumn{3}{c}{\textbf{Fake reviews (\fk{})}} \\ 
\cmidrule(lr){2-4} \cmidrule(lr){5-7}
& \textbf{MC@20 $\uparrow$} & \textbf{RBO@20 $\uparrow$} & \textbf{Recall@20 $\uparrow$} & \textbf{MC@20 $\uparrow$} & \textbf{RBO@20 $\uparrow$} & \textbf{Recall@20 $\uparrow$} \\ 
\midrule
\multicolumn{7}{c}{\lama{}} \\ 
\midrule
\ig{} & 3.34 & 0.20 & 0.73 & 5.31 & 0.31 & 0.70 \\
\shap{} & 2.69 & 0.17 & 0.58 & 3.73 & 0.24 & 0.50 \\
\pca{} & 4.09 & 0.20 & 0.89 & 6.94 & 0.36 & 0.92 \\
\method{}$^{\dmd{}-amp}$ & 4.17 & 0.22 & 0.91 & 7.00 & 3.43 & 0.93 \\ 
\midrule

\multicolumn{7}{c}{\qwn{}} \\ 
\midrule
\ig{} & 3.64 & 0.22 & 0.78 & 4.88 & 0.30 & 0.65 \\
\shap{} & 2.75 & 0.17 & 0.59 & 4.06 & 0.25 & 0.54 \\
\pca{} & 4.11 & 0.21 & 0.89 & 6.88 & 0.35 & 0.92 \\
\method{}$^{\dmd{}-amp}$ & 4.20 & 0.21 & 0.90 & 6.94 & 0.34 & 0.93 \\
\midrule

\multicolumn{7}{c}{\mist{}} \\ 
\midrule
\ig{} & 3.34 & 0.21 & 0.72 & 5.12 & 0.31 & 0.69 \\
\shap{} & 3.12 & 0.20 & 0.67 & 4.23 & 0.26 & 0.57 \\
\pca{} & 4.19 & 0.25 & 0.90 & 6.89 & 0.37 & 0.92 \\
\method{}$^{\dmd{}-amp}$ & 4.15 & 0.23 & 0.90 & 6.93 & 0.34 & 0.93 \\
\bottomrule
\end{tabularx}
\caption{Results for \hx{} and \fk{} for top-$20$ ranked tokens.}
\label{hx_fk_top20}
\end{table*}

\subsection{Selecting top-$7$ modes}
We examine the effect of mode subspace dimensionality by projecting token representations onto the top-$7$ modes compared to top-$5$ in our original setup. As reported in Table \ref{mode_ablation}, selecting additional modes leads to a noticeable drop in performance metrics for \method{}. This performance decrease likely stems from noisy higher-order modes that introduce non-label-intensive tokens into the top attributions. In contrast, \pca{} shows slight, though statistically marginal, performance gains when the number of modes is increased.

\begin{table}[H]
\centering
\scriptsize
\setlength{\tabcolsep}{2pt}
\begin{tabularx}{\columnwidth}{l *{3}{>{\centering\arraybackslash}X}}
\toprule
\textbf{Method} & \textbf{MC} & \textbf{RBO} & \textbf{Rec@k} \\
\midrule
\multicolumn{4}{c}{\textbf{\lama{} -- Hateful (\hx{})}} \\
\midrule
\method{}$^{\dmd{}-amp}$ & 2.79 & 0.25 & 0.64 \\
PCA                      & 2.71 & 0.25 & 0.62 \\
\midrule
\multicolumn{4}{c}{\textbf{\lama{} -- Fake (\fk{})}} \\
\midrule
\method{}$^{\dmd{}-amp}$ & 5.36 & 0.41 & 0.74 \\
PCA                      & 5.11 & 0.41 & 0.70 \\
\midrule
\multicolumn{4}{c}{\textbf{\qwn{} -- Hateful (\hx{})}} \\
\midrule
\method{}$^{\dmd{}-amp}$ & 2.85 & 0.26 & 0.64 \\
PCA                      & 2.80 & 0.26 & 0.63 \\
\midrule
\multicolumn{4}{c}{\textbf{\qwn{} -- Fake (\fk{})}} \\
\midrule
\method{}$^{\dmd{}-amp}$ & 5.33 & 0.41 & 0.74 \\
PCA                      & 4.92 & 0.41 & 0.68 \\
\midrule
\multicolumn{4}{c}{\textbf{\mist{} -- Hateful (\hx{})}} \\
\midrule
\method{}$^{\dmd{}-amp}$ & 2.81 & 0.26 & 0.64 \\
PCA                      & 2.88 & 0.29 & 0.66 \\
\midrule
\multicolumn{4}{c}{\textbf{\mist{} -- Fake (\fk{})}} \\
\midrule
\method{}$^{\dmd{}-amp}$ & 5.17 & 0.40 & 0.72 \\
PCA                      & 4.87 & 0.41 & 0.68 \\
\bottomrule
\end{tabularx}
\caption{Comparative evaluation results for \hx{} and \fk{} datasets, with threshold $\tau=0.5$.}
\label{tab:fakeddit_hatexplain_05}
\end{table}

\begin{table*}[t]
\centering
\scriptsize
\begin{tabularx}{\textwidth}{l|*{3}{C|}|*{3}{C|}}
\toprule
\multirow{2}{*}{\textbf{Method}} & \multicolumn{3}{c|}{\textbf{Negative sentiment}} & \multicolumn{3}{c}{\textbf{Positive sentiment}} \\ 
\cmidrule(lr){2-4} \cmidrule(lr){5-7}
& \textbf{MC@20 $\uparrow$} & \textbf{RBO@20 $\uparrow$} & \textbf{Recall@20 $\uparrow$} & \textbf{MC@20 $\uparrow$} & \textbf{RBO@20 $\uparrow$} & \textbf{Recall@20 $\uparrow$} \\ 
\midrule
\multicolumn{7}{c}{\lama{}} \\ 
\midrule
\pca{} & 5.28 & 0.24 & 0.66 & 5.99 & 0.27 & 0.73 \\
\method{}$^{\hdmd{}-avgamp}$ & 5.21 & 0.23 & 0.66 & 6.04 & 0.26 & 0.74 \\ 
\midrule

\multicolumn{7}{c}{\qwn{}} \\ 
\midrule
\pca{} & 5.46 & 0.25 & 0.68 & 6.08 & 0.28 & 0.75 \\
\method{}$^{\hdmd{}-avgamp}$ & 5.34 & 0.24 & 0.67 & 6.10 & 0.27 & 0.75 \\
\midrule

\multicolumn{7}{c}{\mist{}} \\ 
\midrule
\pca{} & 5.27 & 0.24 & 0.66 & 5.76 & 0.25 & 0.71 \\
\method{}$^{\hdmd{}-avgamp}$ & 5.18 & 0.23 & 0.65 & 5.88 & 0.25 & 0.72 \\
\bottomrule
\end{tabularx}
\caption{Selecting top-7 modes for calculating the attribution for sentiment analysis dataset.}
\label{mode_ablation}
\end{table*}

\section{Label-intensive token alignment vs. decision faithfulness}
To systematically evaluate the semantic relevance of identified tokens, we employ GPT-4.1 to annotate input sequences with label-oriented ground-truth rationale tokens. For any LLM fine-tuned on downstream classification, these label-intensive tokens intuitively represent the core semantic anchors driving the model’s target class selection. Our empirical findings demonstrate that \dmd{} successfully isolates the low-dimensional latent subspaces onto which these label-intensive tokens yield high projection values, revealing their prominent role within the model's implicit representation dynamics. Crucially, we distinguish between \textit{semantic coverage} and \textit{exclusive causal dependency}. Identifying these dominant tokens does not imply that the model relies solely on them to synthesize its output, nor does it suggest that non-labeled contextual tokens are irrelevant to the decision trajectory. As evidenced by our fidelity-by-masking experiments (Section \ref{fidelity})—particularly on the \hx{} dataset where gradient-based baselines like \ig{} exhibit sharper accuracy and confidence drops—\dmd{} prioritizes broad, contextually rich label-intensive patterns over narrow, isolated toxic keywords. Consequently, while \ig{} aggressively degrades task performance by removing single high-leverage tokens, our approach captures a more comprehensive, contextually aligned feature set that reflects the broader latent mechanisms governing sequence processing.

\end{document}